\documentclass{article}

\usepackage[main, final]{neurips_2026}

\usepackage[utf8]{inputenc} 
\usepackage[T1]{fontenc}    
\usepackage{url}            
\usepackage{booktabs}       
\usepackage{amsfonts}       
\usepackage{nicefrac}       
\usepackage{microtype}      
\usepackage{xcolor}         

\usepackage{xspace}
\usepackage{color}
\usepackage{amssymb}
\usepackage{amsmath}
\usepackage{multirow}
\usepackage{colortbl}
\usepackage{makecell,array}
\usepackage{enumitem}
\usepackage{graphicx}
\usepackage{caption}
\usepackage{wrapfig}
\usepackage{fontawesome5}

\usepackage{subcaption}
\definecolor{cvprblue}{rgb}{0.21,0.49,0.74}
\definecolor{tablegray}{RGB}{242,242,242}

\definecolor{tableblue}{RGB}{232,244,252}
\definecolor{tablepeach}{RGB}{255,239,247}

\definecolor{tablepurple}{RGB}{239,239,251}
\definecolor{tablered}{RGB}{249,231,232}

\definecolor{textblue}{RGB}{54,112,164}
\definecolor{textpeach}{RGB}{187,119,150}
\definecolor{textgray}{RGB}{127,127,127}

\newcommand{\gainblue}[1]{$_{\textcolor{textblue}{\tiny \uparrow #1}}$}
\newcommand{\gainpeach}[1]{$_{\textcolor{textpeach}{\tiny \uparrow #1}}$}
\newcommand{\gaingray}[1]{$_{\textcolor{textgray}{\tiny \uparrow #1}}$}
\newcommand{\downgray}[1]{$_{\textcolor{textgray}{\tiny \downarrow #1}}$}

\newcommand{\stdblue}[1]{$_{\textcolor{textblue}{\tiny \pm #1}}$}
\newcommand{\stdpeach}[1]{$_{\textcolor{textpeach}{\tiny \pm #1}}$}
\newcommand{\stdgray}[1]{$_{\textcolor{textgray}{\tiny \pm #1}}$}

\newcommand{\pcolor}[2]{%
  \raisebox{1.00ex}{%
    \makebox[0pt][l]{\textcolor{#1}{\scriptsize #2}}%
  }%
}

\newcommand{\stdcolor}[2]{%
  \raisebox{-0.35ex}{%
    \textcolor{#1}{\scriptsize $\pm$#2}%
  }%
}

\newcommand{\statcellblue}[3]{%
  #1\hspace{0.20em}\pcolor{textblue}{#2}\stdcolor{textblue}{#3}%
}

\newcommand{\statcellgray}[3]{%
  #1\hspace{0.20em}\pcolor{textgray}{#2}\stdcolor{textgray}{#3}%
}

\usepackage[colorlinks=true,
            citecolor=cvprblue,
            linkcolor=cvprblue,
            urlcolor=cvprblue]{hyperref}

\newcommand{\method}{\texttt{VolumeProbe}\xspace}
\newcommand{\lite}{\texttt{Lite}\xspace}
\newcommand{\feat}{\texttt{Feat}\xspace}

\setlist[itemize]{topsep=-1pt, itemsep=1pt, parsep=0pt, partopsep=0pt,left=1pt}

\newcommand{\linkitem}[3]{%
  {\hypersetup{hidelinks}%
   \href{#1}{\mbox{%
     \raisebox{-0.15ex}{%
       \resizebox{!}{9.5pt}{\textcolor{black}{#2}}%
     }%
     \hspace{0.35em}%
     {\fontsize{9.5pt}{12pt}\selectfont
      \textbf{\textcolor{black}{#3}}}%
   }}}%
}

\title{Where to Look Matters: Rethinking Sub-Volume Sampling in 3D Medical Self-supervised Learning}

\author{%
  Junkai Liu \;\;\;\; 
  Le Zhang$^{\dagger}$ \\[2pt]
  School of Engineering, University of Birmingham \\[2pt]
  \texttt{jxl1920@student.bham.ac.uk} \;\;\;\; \texttt{l.zhang.16@bham.ac.uk}\\[7pt]
  \linkitem{https://volumeprobe.github.io/}{\faGlobe}{Website}\qquad
  \linkitem{https://github.com/JK-Liu7/VolumeProbe}{\faGithub}{Code}
}

\begin{document}

\maketitle

{\let\thefootnote\relax\footnotetext{$\dagger$: Corresponding author}}

\begin{abstract}
Self-supervised learning on 3D medical images commonly relies on random sub-volume sampling during pretraining. However, random cropping implicitly treats all spatial regions as equally worth observing, despite anatomical priors and spatial redundancy that make regions inherently unequal in their value for representation learning. To this end, we first show through controlled crop-quality manipulations that better sub-volumes lead to better downstream transfer, revealing the observation policy as an overlooked bottleneck in 3D medical SSL. Motivated by this finding, we revisit sub-volume sampling as a \emph{where-to-look} problem. We propose \textbf{\method}, a plug-and-play framework that replaces random cropping with an offline coarse-to-fine probe bank guided by \textit{anatomy-awareness}, \textit{informativeness}, and \textit{diversity}. We instantiate \method as a lightweight \lite variant using geometric and image-statistical cues, and a diagnostic \feat variant using frozen visual features, to examine whether lightweight volume-intrinsic priors suffice or richer semantic cues reliably provide additional value. Extensive experiments across multiple 3D medical SSL backbones and downstream tasks demonstrate that \method-\lite consistently improves transfer performance over random sampling. Mechanistic analyses further link these gains to more anatomically plausible and informative observations, reduced spatial redundancy, more efficient learning dynamics, and more structured representations. 


\end{abstract}    
\section{Introduction}
\begin{figure*}
\centering
\includegraphics[width=0.90\linewidth]{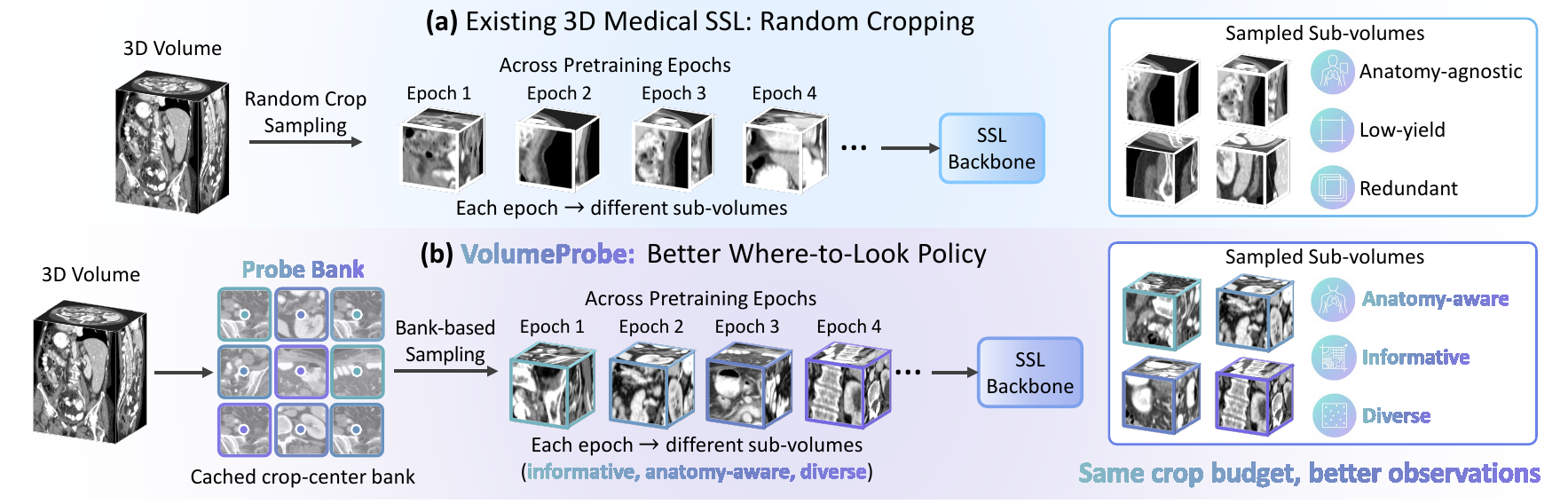}
\vspace{-6pt}
\caption{\textbf{\method reallocates the fixed crop budget toward better observations.} \textbf{(a)} Random cropping in 3D medical SSL often samples low-yield, anatomy-agnostic, and redundant sub-volumes. \textbf{(b)} \method builds a cached probe bank guided by anatomy-awareness, informativeness, and diversity, and samples higher-quality sub-volumes across epochs.}
\vspace{-16pt}
\label{fig:teaser}
\end{figure*}

Self-supervised learning (SSL) has emerged as a powerful paradigm for pretraining on large-scale 3D medical images without dense annotations~\cite{Wu_2024_CVPR, 11274411, Wald_2025_CVPR}. Existing 3D medical SSL pipelines commonly draw fixed-size sub-volumes from each scan via random cropping, which has become the default observation scheme~\cite{hatamizadeh2021swin, wang2023swinmm, 10977020, 11004165}. Representation learning is then driven by repeatedly sampling such sub-volumes over each training iteration. Yet this practice implicitly assumes that all spatial regions are equally valuable for representation learning. In 3D volumes, even after standard foreground cropping, random sub-volume sampling can still allocate the fixed crop budget to low-yield, boundary-dominated, or repetitive regions, while anatomically salient regions provide disproportionately richer observations for representation learning (Fig.~\ref{fig:teaser}(a))~\cite{lee2025hu, tang2024hyspark, Zhou_2021_ICCV}. These observations raise a fundamental question that has been largely underexplored in 3D medical SSL:

\vspace{-7pt}
\begin{quote}
\textit{Given a fixed crop budget, where should a 3D medical SSL model look?}
\end{quote}
\vspace{-7pt}

Our \textbf{core insight} is that, under the sub-crop setting, the quality of sampled sub-volumes can directly shape representation learning in 3D medical SSL. Pretraining is ultimately driven by the evidence exposed through repeated observations~\cite{Jiang_2023_ICCV, TANG2026103770, Hou_2024_CVPR}. Intuitively, crops covering anatomically plausible and information-rich regions provide meaningful supervisory signals, whereas redundant or low-yield crops squander the limited observation budget~\cite{Taher_2024_CVPR, qi2025fdas, hosseinzadeh2025learning, Wald_2025_ICCV}. As we illustrate in Sec.~\ref{sec:motivation}, these differences in crop quality already translate into measurable differences in downstream performance. This suggests that random cropping should not be viewed merely as a preprocessing convenience, but rather as a critical observation policy. Building on this insight, we reformulate sub-volume sampling as a structured \emph{where-to-look} problem rather than an unconstrained random process, enabling the sub-volume cropping budget to be reallocated toward better observations (Fig.~\ref{fig:teaser}(b)).

These observations motivate us to propose \method, a training-free, plug-and-play framework for structured sub-volume sampling in 3D medical SSL. \method replaces random cropping with a probe bank built on three simple and intuitive principles: (i) \emph{anatomy-awareness}, to favor anatomically plausible regions; (ii) \emph{informativeness}, to prioritize sub-volumes with richer structural content; and (iii) \emph{diversity}, to reduce redundancy among observations. Concretely, \method scores candidate locations, i.e., 3D crop-center coordinates in the normalized volume space, under these principles to build a cached probe bank offline, from which sub-volumes are drawn during pretraining. To understand what makes an effective observation policy under this setting, we instantiate \method in two variants: a lightweight \lite version driven by low-cost geometric priors, and a \feat version built on frozen dense features. This lets us ask whether lightweight priors already suffice, or whether richer semantic cues offer additional value. Across both variants, \method reallocates a fixed crop budget toward higher-quality observations without modifying the SSL backbone or training objective. Our main contributions can be summarized as follows:
\begin{itemize}
    \item \textbf{Finding.} Through a controlled motivation study, we unveil sub-volume observation quality as a previously underexplored factor that influences representation learning in 3D medical SSL.

    \item \textbf{Framework.} We propose \method, a training-free, plug-and-play module that instantiates the \emph{where-to-look} formulation through a probe bank in both \lite and \feat variants.

    \item \textbf{Understanding.} Beyond benchmark gains, principle-driven mechanistic analyses and the \lite vs \feat comparison reveal how observation quality translates into learning and clarify when lightweight priors suffice versus when richer semantic cues yield additional value.

    \item \textbf{Broad Applicability.} Extensive experiments on 3 SSL frameworks across 3 downstream tasks show that \method consistently improves transfer performance over random cropping, demonstrating its plug-and-play compatibility with diverse SSL backbones.
\end{itemize}

\section{Motivation Analysis: Why Observation Quality Matters}
\label{sec:motivation}
In this section, we investigate whether sub-volume observation quality is relevant to representation learning in 3D medical SSL through a small-scale experiment using VoCo~\cite{Wu_2024_CVPR} as the SSL backbone.

\begin{figure*}
\centering
\includegraphics[width=\linewidth]{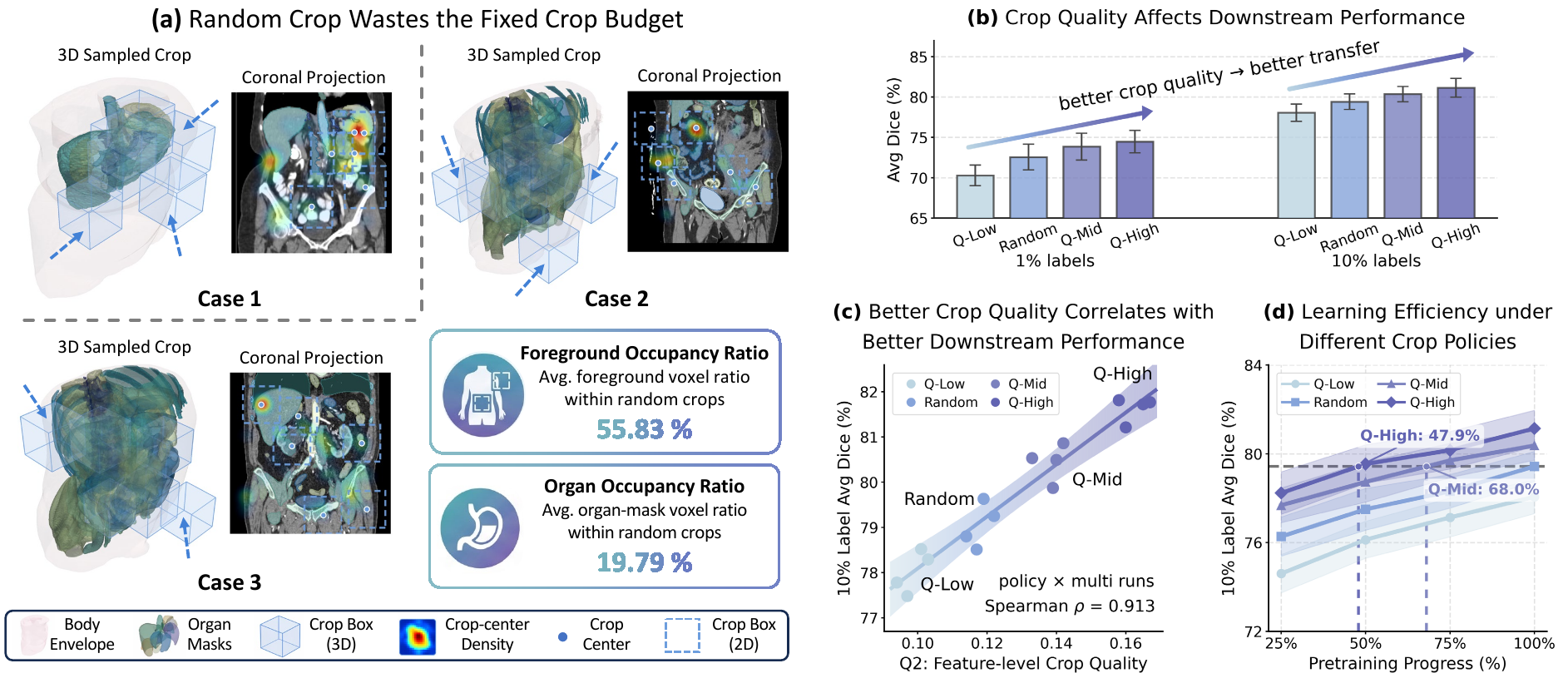}
\vspace{-14pt}
\caption{\textbf{Crop quality matters for 3D medical representation learning.} \textbf{(a)} Random cropping frequently spends the fixed crop budget on structure-sparse regions. \textbf{(b)} Policies defined by a semantic quality axis $Q_1$ yield a monotonic change in downstream Dice. \textbf{(c)} This trend remains under an alternative feature-level proxy $Q_2$. \textbf{(d)} Better crop quality further improves learning efficiency.}
\vspace{-16pt}
\label{fig:motivation}
\end{figure*}

\textbf{Observation waste under a fixed crop budget.} As shown in Fig.~\ref{fig:motivation}(a), even after standard foreground cropping, random crops can still cover peripheral or structure-sparse regions. Quantitatively, they contain 55.83\% foreground voxels on average, yet only 19.79\% overlap annotated organ masks on the mask-available subset~\cite{landman2015miccai, SETIO20171, wasserthal2023totalsegmentator, ma2024automatic}. This gap indicates that random sampling often observes non-background but anatomy-agnostic regions, leaving organ-rich evidence underutilized.

\textbf{Sub-volume quality shifts downstream transfer.} To test whether this waste affects representation learning, we construct four static probe policies along an oracle semantic quality axis $Q_1$, defined as the number of distinct organ classes contained in each crop. For each mask-available volume, we sample candidate crop centers, rank them by $Q_1$, and form static banks from the bottom (\textit{Q-Low}), middle (\textit{Q-Mid}), and top (\textit{Q-High}) quartiles, while \textit{Random} samples uniformly from all candidates. Pretraining follows the VoCo-10k protocol~\cite{Wu_2024_CVPR} for 50\% of the full training steps, with $Q$-manipulation applied only to volumes with available organ masks~\cite{landman2015miccai, SETIO20171, wasserthal2023totalsegmentator, ma2024automatic}. Fig.~\ref{fig:motivation}(b) shows that downstream average Dice on BTCV~\cite{landman2015miccai} and AMOS~\cite{NEURIPS2022_ee604e1b} increases monotonically from \textit{Q-Low} to \textit{Q-High} under both 1\% and 10\% label regimes, with \textit{Random} lying in between. This suggests that crop quality is a controllable axis that random sampling does not fully exploit.

\textbf{The quality-transfer link holds under an alternative proxy.} 
Fig.~\ref{fig:motivation}(c) replaces $Q_1$ with a feature-level 
quality $Q_2$ computed as the variance of frozen MedicalNet~\cite{chen2019med3d} features on the sampled crops. The correlation with downstream average Dice under 10\% labels remains positive (Spearman $\rho=0.913$), so the quality-transfer relationship is not bound to a specific oracle.

\textbf{Crop quality improves learning efficiency.} Fig.~\ref{fig:motivation}(d) reveals a clear ordering throughout pretraining, with \textit{Q-High} and \textit{Q-Mid} consistently above \textit{Random} and \textit{Q-Low} lowest. Notably, \textit{Q-High} reaches the final Dice of Random@100\% at only 47.9\% pretraining progress, and \textit{Q-Mid} at 68.0\%, showing that higher-quality observations yield comparable transfer with less pretraining.

However, neither $Q_1$ nor $Q_2$ is available during pretraining, as the former requires organ labels and the latter an external feature extractor. These findings thus motivate \method as a training-free observation policy that approximates the quality axis using only volume-intrinsic signals.

\section{Methodology}

\subsection{Overview}
\label{sec:overview}
\begin{figure*}
\centering
\includegraphics[width=1.00\linewidth]{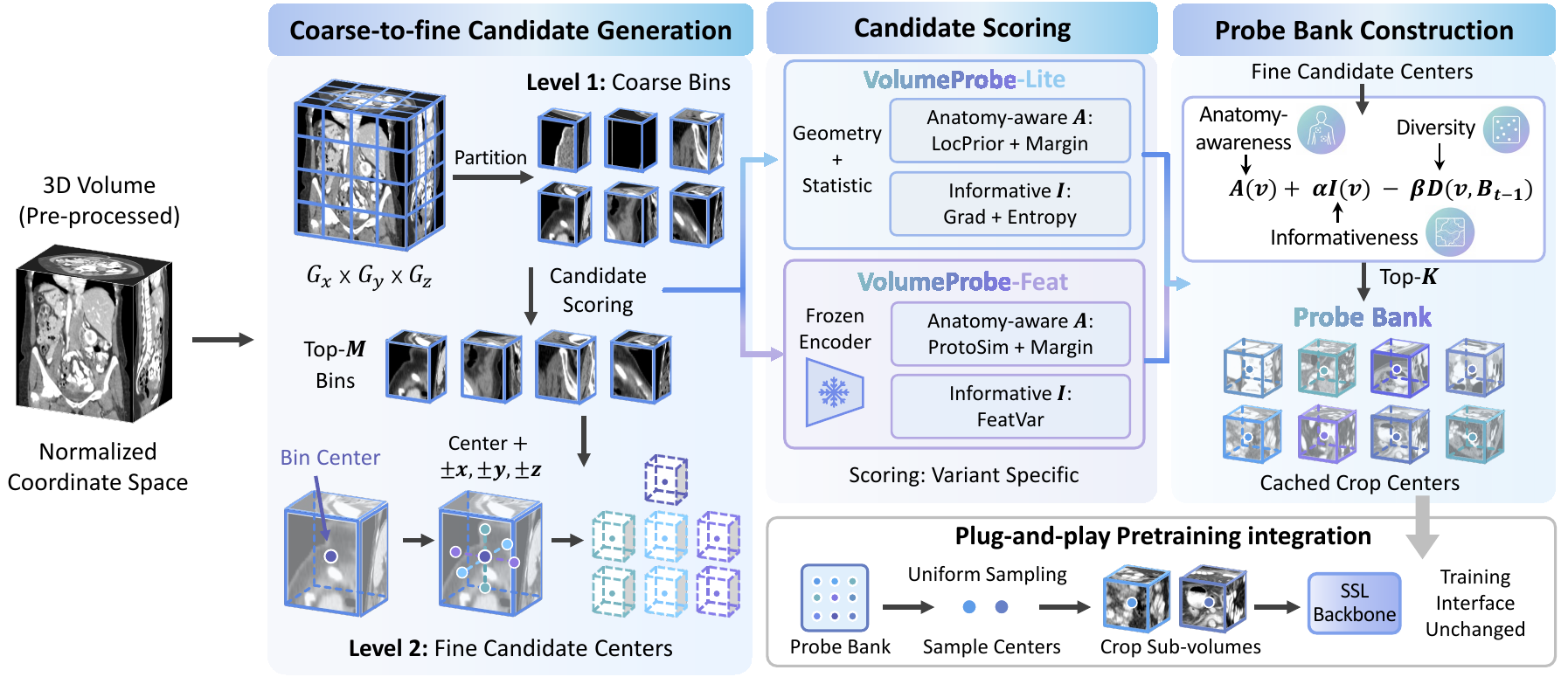}
\vspace{-12pt}
\caption{\textbf{Overview of \method.} \method constructs an offline probe bank of normalized 3D crop centers guided by anatomy-awareness, informativeness, and diversity. Pretraining then samples sub-volumes from this bank without changing the SSL backbone or objective.}
\vspace{-16pt}
\label{fig:framework}
\end{figure*}

Let $\mathbf{x}$ denote a preprocessed 3D volume with normalized coordinate space $\Omega = [0,1]^3$. As shown in Fig.~\ref{fig:framework}, \method\ first builds a coarse-to-fine candidate set $\mathcal{V}(\mathbf{x}) \subset \Omega$, where Level~1 partitions $\Omega$ into a coarse grid of $G_x \times G_y \times G_z$ bins and Level~2 refines only the selected high-value bins into fine candidate centers (i.e., 3D position coordinates). From this pool, \method\ selects a probe bank $\mathcal{B}(\mathbf{x})=\{v_1,\dots,v_K\}$ of $K$ sub-volume centers offline, from which one center is sampled during pretraining to crop a fixed-size sub-volume. In what follows, we establish three principles for candidate scoring and probe bank construction (Sec.~\ref{sec:principles}), instantiate them in a \lite variant (Sec.~\ref{sec:lite}) and a \feat variant (Sec.~\ref{sec:feat}), and describe pretraining integration (Sec.~\ref{sec:pretraining_integration}).

\subsection{Principles for Probe Bank Construction}
\label{sec:principles}
An effective probe bank should concentrate the fixed crop budget on candidates that are worth observing. We characterize each candidate $v$ through three principles. The \textit{anatomy-aware} term $A(v)$ favors locations in meaningful body regions rather than weak or boundary-dominated areas. The \textit{informativeness} term $I(v)$ rewards candidates whose local region contains richer structural content. The \textit{diversity} term, expressed as a redundancy penalty $D(v,\mathcal{B})$, discourages candidates that would make the bank locally overcrowded. We combine them into a unified objective as
\begin{equation}
v_t=\arg\max_{v\in \mathcal{V}(x)\setminus \mathcal{B}_{t-1}}
\Big[A(v)+\alpha I(v)-\beta D(v,\mathcal{B}_{t-1})\Big],\qquad t=1,\dots,K.
\label{eqn:objective}
\end{equation}
where $\alpha$ and $\beta$ are weight parameters. The resulting greedy construction is tractable and deterministic, and \method-\lite and \method-\feat share this formulation, differing only in how the three terms are quantified.

\subsection{\method-\lite}
\label{sec:lite}
\method-\lite instantiates probe quality using geometric priors and image statistics. We rank Level-1 bins by $A(v)+\alpha I(v)$ in Eq.~\ref{eqn:objective} and keep the top-$M$ for refinement. Within each retained bin, Level-2 candidates consist of the bin center and up to six axis-aligned neighbors offset by a fixed fraction of the bin size along $\pm x, \pm y, \pm z$, clamped to stay inside the bin. We then rescore these fine candidates and construct the top-$K$ probe bank by greedy redundancy-aware selection.


\textbf{Anatomy-aware.}
We model anatomical plausibility using a simple geometric prior:
\begin{equation}
A(v)=\mathrm{LocPrior}(v) + \mathrm{Margin}(v).
\end{equation}
Here, $\mathrm{LocPrior}(v) = 1-\mathrm{dist}$, where $\mathrm{dist}$ is the normalized distance from $v$ to the volume center with a smaller weight on the $z$ axis to reflect volumetric anisotropy. $\mathrm{Margin}(v)$ is the normalized minimum distance from $v$ to the volume boundary, penalizing candidates near the border. This geometric prior is not a center-cropping rule, but an anatomy-correlated plausibility cue that is jointly combined with informativeness, diversity, and coarse-to-fine candidate selection.

\textbf{Informative.}
We quantify local informativeness by combining gradient and entropy cues:
\begin{equation}
I(v)=\,\widetilde{\mathrm{Grad}}(v) + \,\widetilde{\mathrm{Ent}}(v).
\end{equation}
where $\widetilde{\mathrm{Grad}}(v)$ is the normalized mean gradient magnitude and $\widetilde{\mathrm{Ent}}(v)$ is the normalized intensity entropy computed in a local evaluation patch around $v$. Coarse-bin scoring evaluates the two terms over the entire bin, whereas fine-candidate scoring uses a local patch of the pretraining sub-crop size centered at $v$. The gradient term favors stronger structural transitions, while the entropy term avoids overly homogeneous regions.

\textbf{Diversity.}
To prevent the bank from collapsing to a few nearby high-score locations, we use an axis-wise redundancy penalty:
\begin{equation}
D(v,B)=\max_{u\in B}\prod_{a\in\{x,y,z\}}\left[1-\frac{|v_a-u_a|}{\delta_a}\right]_+ .
\end{equation}
where $\delta_a = \eta \cdot \mathrm{roi}_a / \mathrm{shape}_a$ matches the anisotropic footprint of the pretraining crop and $\eta$ is a fixed factor. The product form makes the penalty large only when $v$ is close to an existing bank member along all three axes simultaneously, matching the overlap pattern of axis-aligned 3D crops.

Relying on geometric properties and statistics alone, \method-\lite serves as a controlled reference point against which the benefit of richer semantic scoring (Sec.~\ref{sec:feat}) can be isolated.


\subsection{\method-\feat}
\label{sec:feat}
\method-\feat investigates whether frozen semantic features provide additional value beyond the statistical cues in \method-\lite. It shares the same candidate generation and bank-selection pipeline, but replaces the scoring cues with probe-level features extracted by a frozen vision encoder.

\textbf{Anatomy-aware.}
We replace the geometric anatomy prior with a feature-consistency prior. Let $\bar{\mathbf{f}}_v$ be the pooled frozen feature within the local region centered at $v$, and let $b(v)$ denote its coarse bin. For each bin $b$, we compute a prototype $\boldsymbol{\mu}_b$ by averaging probe-level features assigned to it across the training set. The anatomy-aware score is
\begin{equation}
A_{\mathrm{feat}}(v)
=
\widetilde{\mathrm{ProtoSim}}(v)
+
\widetilde{\mathrm{Margin}}(v),
\quad
\mathrm{ProtoSim}(v)
=
\frac{1+\cos(\bar{\mathbf{f}}_v,\boldsymbol{\mu}_{b(v)})}{2}.
\end{equation}
where $\mathrm{ProtoSim}(v)$ measures whether a probe is feature-consistent with its coarse anatomical region, while $\mathrm{Margin}(v)$ reuses the boundary suppression from \method-\lite. Both terms are normalized to $[0,1]$ per volume, yielding $\widetilde{\mathrm{ProtoSim}}(v)$ and $\widetilde{\mathrm{Margin}}(v)$, respectively.

\textbf{Informative.}
We quantify informativeness as local heterogeneity in the frozen feature space:
\begin{equation}
I_{\mathrm{feat}}(v)
=
\widetilde{\mathrm{FeatVar}}(v),
\quad
\mathrm{FeatVar}(v)
=
\frac{1}{|R(v)|}
\sum_{p\in R(v)}
\left\|
\mathbf{f}_p-\bar{\mathbf{f}}_v
\right\|_2^2.
\end{equation}
where $R(v)$ is the local probe region and $\mathbf{f}_p$ is the frozen feature at position $p$.

\textbf{Diversity.}
We keep the axis-wise redundancy penalty unchanged from \method-\lite, so the \lite-vs-\feat comparison isolates the effect of the scoring source while using the same candidate construction and bank-selection interface.

\subsection{Pretraining Integration}
\label{sec:pretraining_integration}
Once the per-volume probe bank $\mathcal{B}(\mathbf{x})$ is constructed offline, \method\ integrates into SSL pretraining without learnable parameters, online reweighting, or substantial recurring overhead. During pretraining, we sample crop centers either from the cached bank or via standard random cropping, then pass the resulting fixed-size sub-volumes to the unchanged SSL backbone.
\section{Experiments}
We organize our experiments around five research questions, covering both downstream performance and the mechanisms behind the observed gains.
\begin{itemize}
    \item \textbf{RQ1}: Does \method improve medical SSL across downstream datasets and SSL backbones?
    \item \textbf{RQ2}: Do lightweight geometric priors (\lite) already suffice for effective sub-volume probing, or does richer semantic information (\feat) offer additional value?
    \item \textbf{RQ3}: Are the three design principles of \method (anatomy-awareness, informativeness, and diversity) necessary for its downstream gains?
    \item \textbf{RQ4}: How do the anatomy-aware, informative, and diversity principles shape the resulting probe bank and probing behavior?
    \item \textbf{RQ5}: Do these improved probing behaviors translate into measurable changes in how the pretraining model learns?
        
\end{itemize}

\subsection{Experimental Setup}
\label{sec:exp_set}
\textbf{Datasets.}
We integrate \method into three 3D medical SSL frameworks, VoCo~\cite{Wu_2024_CVPR}, S2DC~\cite{Pan_2025_ICCV}, and MedGMAE~\cite{fu2026medgmae}, following their original pretraining protocols~\cite{landman2015miccai, clark2013cancer, SETIO20171, doi:10.1148/radiol.2021210384, ma2024automatic, grossberg2018imaging, wasserthal2023totalsegmentator, LI2024103285}, and replacing only the random cropping with \method while keeping the number of sampled sub-volumes unchanged. Downstream evaluation spans three tasks. For segmentation, we use BTCV~\cite{landman2015miccai}, AMOS~\cite{NEURIPS2022_ee604e1b}, WORD~\cite{LUO2022102642}, and Atlas-MRI~\cite{data8050079} under 1\%, 10\%, and 100\% data proportions with official splits. For classification, we adopt CC-CCII~\cite{zhang2020clinically}. For registration, we evaluate on IXI~\cite{KIM2021102036} and OASIS~\cite{6795067} following~\cite{Wu_2024_CVPR, 11274411}. Additional details are deferred to Appendix~\ref{app:data}.

\textbf{Implementation Details.}
For our 3D medical SSL baselines~\cite{Wu_2024_CVPR, Pan_2025_ICCV, fu2026medgmae}, we strictly follow their original default settings for pretraining and downstream evaluation, including preprocessing, crop size, number of sampled sub-volumes, backbone, and task-specific protocols. For \method, we use the coarse grid $(G_x, G_y, G_z) = (4, 4, 2)$, with $M=24$ and $K=40$. The weights $\alpha$ and $\beta$ are both set to $1.0$ and the redundancy scale factor $\eta$ to $0.35$. For \method-\feat, we use 3DINO-ViT~\cite{xu2025generalizable} as the feature extractor. Further details are provided in Appendix~\ref{app:imple}.

\subsection{Experiments on Downstream Tasks (RQ1)}
We compare \method against both General~\cite{Chen_2023_WACV, Xie_2022_CVPR, He_2020_CVPR} and Medical SSL methods~\cite{Hatamizadeh_2022_WACV, hatamizadeh2021swin, He_2023_CVPR, 11004165, 10977020, Wu_2024_CVPR, Pan_2025_ICCV, fu2026medgmae}, using official implementations and pretrained weights where available, or results drawn from prior work~\cite{Wu_2024_CVPR, fu2026medgmae}. As shown in Table~\ref{tab:seg_cls}, substituting random cropping with \method yields gains across all three SSL backbones on four datasets, with the most pronounced improvements under low-label regimes. Beyond segmentation, \method further delivers steady accuracy improvements on CC-CCII~\cite{zhang2020clinically}, demonstrating that the representational benefits generalize to classification. Table~\ref{tab:reg} further corroborates these findings on registration, where gains on IXI~\cite{KIM2021102036} and OASIS~\cite{6795067} are modest yet consistently positive across all three backbones, confirming the plug-and-play effectiveness of \method across diverse downstream tasks. 

\begin{table*}[t]
\centering
\setlength{\tabcolsep}{1.5mm}
\caption{\textbf{Results on segmentation and classification tasks.} 
For segmentation, we report Dice Score (\%) under different training data proportions. For classification, we report Accuracy (\%). The best results are \textbf{bolded}, and the second best results are \underline{underlined}.}
\vspace{-5pt}
\renewcommand\arraystretch{0.95}
\resizebox{1.0\linewidth}{!}{
\begin{tabular}{lccc|ccc|ccc|ccc|c}
\toprule
\multirow{3}{*}{\textbf{Method}}
& \multicolumn{12}{c|}{\textbf{Dice Score (\%) $\uparrow$}}
& \textbf{Acc. (\%) $\uparrow$} \\
\cmidrule(lr){2-13}\cmidrule(lr){14-14}
& \multicolumn{3}{c|}{\textbf{BTCV}} 
& \multicolumn{3}{c|}{\textbf{AMOS}} 
& \multicolumn{3}{c|}{\textbf{WORD}} 
& \multicolumn{3}{c|}{\textbf{Atlas-MRI}} 
& \multirow{2}{*}{\textbf{CC-CCII}}\\
\cmidrule(lr){2-4}\cmidrule(lr){5-7}\cmidrule(lr){8-10}\cmidrule(lr){11-13}
& 1\% & 10\% & 100\%
& 1\% & 10\% & 100\%
& 1\% & 10\% & 100\%
& 1\% & 10\% & 100\%
& \\
\midrule
\rowcolor{tablegray!80}
\multicolumn{14}{c}{\textit{From Scratch}} \\
UNETR~\cite{Hatamizadeh_2022_WACV}
  & 28.05 & 42.85 & 79.82
  & 23.67 & 60.06 & 77.02
  & 30.89 & 73.76 & 78.73
  & 18.65 & 40.83 & 49.97
  & 88.65 \\
Swin-UNETR~\cite{hatamizadeh2021swin}
  & 27.71 & 51.33 & 80.53
  & 28.94 & 63.45 & 82.51
  & 43.68 & 74.03 & 80.39
  & 17.19 & 39.46 & 48.50
  & 88.32 \\
\midrule
\rowcolor{tablegray!80}
\multicolumn{14}{c}{\textit{General SSL}} \\
MAE3D~\cite{Chen_2023_WACV}
  & 31.86 & 55.79 & 81.33
  & 37.82 & 62.83 & 82.26
  & 39.23 & 72.53 & 78.97
  & 28.46 & 42.01 & 52.85
  & 90.09\\
SimMIM~\cite{Xie_2022_CVPR}
  & 34.73 & 56.34 & 76.03
  & 30.98 & 63.70 & 80.83
  & 37.56 & 73.25 & 79.60
  & 26.22 & 41.53 & 52.47
  & 91.27 \\
MoCo v3~\cite{He_2020_CVPR}
  & 38.27 & 77.91 & 79.54
  & 34.67 & 65.76 & 81.95
  & 41.43 & 72.89 & 80.73
  & 30.57 & 43.28 & 54.31
  & 87.76 \\
\midrule
\rowcolor{tablegray!80}
\multicolumn{14}{c}{\textit{Medical Image SSL}} \\
Swin-UNETR~\cite{hatamizadeh2021swin}
  & 28.52 & 48.31 & 81.54
  & 24.63 & 65.49 & 80.97
  & 48.16 & 75.07 & 81.79
  & 24.09 & 41.96 & 53.38
  & 89.42 \\
GVSL~\cite{He_2023_CVPR}
  & 24.86 & 41.79 & 81.87
  & 34.39 & 63.45 & 81.38
  & 37.87 & 74.10 & 82.36
  & 28.63 & 42.83 & 54.86
  & 93.26\\
GL-MAE~\cite{11004165}
  & 52.46 & 65.83 & 82.01
  & 48.14 & 72.43 & 84.20
  & 62.60 & 75.42 & 83.17
  & 32.54 & 46.02 & 57.19
  & 92.13 \\
MIM~\cite{10977020}
  & 60.53 & 74.20 & 81.97
  & 50.40 & 73.59 & 84.72
  & 60.58 & 77.03 & 83.98
  & 33.19 & 45.20 & 57.03
  & 92.75 \\
\midrule
VoCo~\cite{Wu_2024_CVPR}
  & 63.33 & 77.85 & 83.85
  & 55.81 & 73.34 & 84.44
  & 64.24 & 77.28 & 84.06
  & 34.85 & 47.56 & 58.83
  & 93.45 \\
\rowcolor{tableblue!80}
\textbf{+ \method-\lite} 
  & 67.21\gainblue{3.88} & \underline{79.39}\gainblue{1.54} & 83.97\gainblue{0.12}
  & 58.64\gainblue{2.83} & 74.95\gainblue{1.61} & \textbf{87.52}\gainblue{3.08}
  & 68.93\gainblue{4.69} & \textbf{80.12}\gainblue{2.84} & \textbf{85.56}\gainblue{1.50}
  & 38.68\gainblue{3.83} & 49.27\gainblue{1.71} & 59.98\gainblue{1.15}
  & 93.98\gainblue{0.53} \\
\rowcolor{tablepeach!70}
\textbf{+ \method-\feat}
  & 68.07\gainpeach{4.74} & \textbf{79.51}\gainpeach{1.66} & 83.89\gainpeach{0.04}
  & 57.86\gainpeach{2.05} & 74.70\gainpeach{1.36} & 87.13\gainpeach{2.69}
  & 68.42\gainpeach{4.18} & 79.74\gainpeach{2.46} & 84.99\gainpeach{0.93}
  & 37.93\gainpeach{3.08} & 49.06\gainpeach{1.50} & 59.42\gainpeach{0.59}
  & 94.25\gainpeach{0.80} \\
\midrule
S2DC~\cite{Pan_2025_ICCV}
  & 66.77 & 77.93 & 84.09
  & 55.19 & 74.28 & 84.68
  & 69.14 & 78.96 & 84.46
  & 33.79 & 46.18 & 58.53
  & 94.19 \\
\rowcolor{tableblue!80}
\textbf{+ \method-\lite} 
  & \textbf{69.47}\gainblue{2.70} & 78.74\gainblue{0.81} & \textbf{84.43}\gainblue{0.34}
  & 58.83\gainblue{3.64} & 75.31\gainblue{1.03} & \underline{87.17}\gainblue{2.49}
  & 69.73\gainblue{0.59} & \underline{80.04}\gainblue{1.08} & 85.22\gainblue{0.76}
  & 37.16\gainblue{3.37} & 48.08\gainblue{1.90} & 59.44\gainblue{0.91}
  & \textbf{94.87}\gainblue{0.68} \\
\rowcolor{tablepeach!70}
\textbf{+ \method-\feat}
  & \underline{68.76}\gainpeach{1.99} & 78.92\gainpeach{0.99} & \underline{84.19}\gainpeach{0.10}
  & 59.47\gainpeach{4.28} & 75.17\gainpeach{0.89} & 86.87\gainpeach{2.19}
  & 70.28\gainpeach{1.14} & 79.82\gainpeach{0.86} & \underline{85.43}\gainpeach{0.97}
  & 38.19\gainpeach{4.40} & 48.83\gainpeach{2.65} & 59.81\gainpeach{1.28}
  & \underline{94.34}\gainpeach{0.15} \\
\midrule
MedGMAE~\cite{fu2026medgmae}
  & 66.19 & 77.11 & 83.22
  & 58.79 & 75.65 & 84.90
  & 68.09 & 77.73 & 84.19
  & 36.24 & 48.16 & 59.42
  & 93.08 \\
\rowcolor{tableblue!80}
\textbf{+ \method-\lite} 
  & 66.85\gainblue{0.66} & 78.44\gainblue{1.33} & 83.48\gainblue{0.26}
  & \textbf{60.26}\gainblue{1.47} & \textbf{77.14}\gainblue{1.49} & 85.62\gainblue{0.72}
  & \underline{70.67}\gainblue{2.58} & 79.42\gainblue{1.69} & 84.36\gainblue{0.17}
  & \textbf{39.71}\gainblue{3.47} & \textbf{50.23}\gainblue{2.07} & \underline{60.19}\gainblue{0.77}
  & 93.58\gainblue{0.50} \\
\rowcolor{tablepeach!70}
\textbf{+ \method-\feat}
  & 66.57\gainpeach{0.38} & 79.09\gainpeach{1.98} & 83.29\gainpeach{0.07}
  & \underline{59.57}\gainpeach{0.78} & \underline{76.80}\gainpeach{1.15} & 85.93\gainpeach{1.03}
  & \textbf{71.15}\gainpeach{3.06} & 79.08\gainpeach{1.35} & 84.25\gainpeach{0.06}
  & \underline{39.48}\gainpeach{3.24} & \underline{49.83}\gainpeach{1.67} & \textbf{60.40}\gainpeach{0.98}
  & 93.81\gainpeach{0.73} \\
\bottomrule
\end{tabular}
}
\label{tab:seg_cls}
\vspace{-14pt}
\end{table*}

\subsection{\lite vs. \feat: Are Lightweight Priors Sufficient? (RQ2)}
Tables~\ref{tab:seg_cls} and~\ref{tab:reg} show that \method-\lite, relying only on training-free geometric and statistical cues, already achieves strong and consistent gains. Meanwhile, incorporating frozen semantic features in \method-\feat does not reliably improve over \method-\lite. This observation indicates that \textbf{volume-intrinsic priors are sufficient to construct effective probe banks}. We therefore treat \method-\feat as a diagnostic external-feature variant rather than the main method. Unless otherwise specified, \method in the remainder of this paper refers to \method-\lite.

\subsection{Ablation Study (RQ3)}
\label{sec:ablation}

\textbf{Component Ablation.}
In Table~\ref{tab:ablation}, \method delivers consistent gains over the random-cropping baseline on both VoCo~\cite{Wu_2024_CVPR} and S2DC~\cite{Pan_2025_ICCV}, while ablating any single principle degrades performance. Notably, dropping anatomy-awareness or informativeness incurs the largest drops, underscoring that anatomically plausible and content-rich observations are particularly essential for robust transfer. Please refer to Appendices~\ref{app:ablation} and~\ref{app:efficiency} for more fine-grained ablation studies and efficiency analysis.

\begin{table*}[t]
\centering
\begin{minipage}[t]{0.42\textwidth}
    \centering
    \caption{\textbf{Results on the registration task on IXI and OASIS datasets}.}
    \label{tab:reg}
    \vspace{-6pt}
    \setlength{\tabcolsep}{2.5mm}
    \renewcommand\arraystretch{1.00}
    \resizebox{0.90\linewidth}{!}{
    \begin{tabular}{lcc}
    \toprule
    \multirow{2}{*}{\textbf{Method}} & \multicolumn{2}{c}{\textbf{Dice Score (\%) $\uparrow$}} \\
    \cmidrule(lr){2-3}
     & \textbf{IXI} & \textbf{OASIS} \\
    \midrule
    \rowcolor{tablegray!80}
    \multicolumn{3}{c}{\textit{From Scratch}} \\
    VoxelMorph~\cite{8633930} & 71.54 & 78.55 \\
    TransMorph~\cite{CHEN2022102615} & \underline{74.46} & 81.61 \\
    Swin-UNETR~\cite{hatamizadeh2021swin} & 72.63 & 81.81 \\
    \midrule
    \rowcolor{tablegray!80}
    \multicolumn{3}{c}{\textit{Medical Image SSL}} \\
    SUP~\cite{Tang_2022_CVPR} & 67.67 & 81.54 \\
    SuPreM~\cite{li2024well} & 72.90 & 81.18 \\
    \midrule
    VoCo~\cite{Wu_2024_CVPR} & 73.66 & 84.38 \\
    \rowcolor{tableblue!80}
    \textbf{+ \method-\lite} & 74.39\gainblue{0.73} & 85.55\gainblue{1.17} \\
    \rowcolor{tablepeach!70}
    \textbf{+ \method-\feat} & 74.17\gainpeach{0.51} & 85.19\gainpeach{0.81} \\
    \midrule
    S2DC~\cite{Pan_2025_ICCV} & 72.93 & 82.97 \\
    \rowcolor{tableblue!80}
    \textbf{+ \method-\lite} & 73.70\gainblue{0.77} & 84.09\gainblue{1.12} \\
    \rowcolor{tablepeach!70}
    \textbf{+ \method-\feat} & 73.98\gainpeach{1.05} & 83.86\gainpeach{0.89} \\
    \midrule
    MedGMAE~\cite{fu2026medgmae} & 73.69 & 85.71 \\
    \rowcolor{tableblue!80}
    \textbf{+ \method-\lite} & \textbf{74.88}\gainblue{1.19} & \underline{85.96}\gainblue{0.25} \\
    \rowcolor{tablepeach!70}
    \textbf{+ \method-\feat} & 74.28\gainpeach{0.59} & \textbf{86.17}\gainpeach{0.46} \\
    \bottomrule
    \end{tabular}
    }
\end{minipage}
\hfill
\begin{minipage}[t]{0.54\textwidth}
    \centering
    \caption{\textbf{Ablation study of our principles for probe selection}.}
    \label{tab:ablation}
    \vspace{-0pt}
    \setlength{\tabcolsep}{1.8mm}
    \renewcommand\arraystretch{1.10}
    \resizebox{1.00\linewidth}{!}{
    \begin{tabular}{lccc}
    \toprule
    \multirow{2}{*}{\textbf{Method}} & \multicolumn{3}{c}{\textbf{Dice Score (\%) $\uparrow$}} \\
    \cmidrule(lr){2-4}
    & \textbf{BTCV} & \textbf{AMOS} & \textbf{WORD} \\
    \midrule
    VoCo~\cite{Wu_2024_CVPR}               & 83.85 & 84.44 & 84.06  \\
    \rowcolor{tableblue!80}
    \textbf{+ \method}      & 83.97\gainblue{0.12} & 87.52\gainblue{3.08} & 85.56\gainblue{1.50}  \\
    w/o $A$              & 83.87\gaingray{0.02} & 84.96\gaingray{0.52} & 84.47\gaingray{0.41} \\
    w/o $I$              & 83.86\gaingray{0.01} & 85.89\gaingray{1.45} & 85.04\gaingray{0.98} \\
    w/o $D$              & 83.93\gaingray{0.08} & 87.10\gaingray{2.66} & 85.29\gaingray{1.23} \\
    \midrule
    S2DC~\cite{Pan_2025_ICCV}               & 84.09 & 84.68 & 84.46  \\
    \rowcolor{tableblue!80}
    \textbf{+ \method}       & 84.43\gainblue{0.34} & 87.17\gainblue{2.49} & 85.22\gainblue{0.76} \\
    w/o $A$              & 84.20\gaingray{0.11} & 86.20\gaingray{1.52} & 84.73\gaingray{0.27}  \\
    w/o $I$              & 84.27\gaingray{0.18} & 85.45\gaingray{0.77} & 84.81\gaingray{0.35}  \\
    w/o $D$              & 84.35\gaingray{0.26} & 86.93\gaingray{2.25} & 85.10\gaingray{0.64}  \\
    \bottomrule
    \end{tabular}
    }
\end{minipage}
\vspace{-8pt}
\end{table*}

\begin{wraptable}{r}{0.55\textwidth}
\centering
\vspace{-2pt}
\caption{\textbf{Comparison of different non-random sampling strategies for VoCo}.}
\setlength{\tabcolsep}{2.0mm}
\renewcommand\arraystretch{1.0}
\resizebox{0.98\linewidth}{!}{
\begin{tabular}{cccc}
\toprule
\multirow{2}{*}{\textbf{Sampling Method}} & \multicolumn{3}{c}{\textbf{Dice Score (\%) $\uparrow$}} \\
\cmidrule(lr){2-4}
& \textbf{BTCV} & \textbf{AMOS} & \textbf{WORD} \\
\midrule
VoCo~\cite{Wu_2024_CVPR} & 83.85 & 84.44 & 84.06 \\
\midrule
+ Center-Biased            & 83.79\downgray{0.06} & 85.32\gaingray{0.88} & 83.95\downgray{0.11}   \\
+ Uniform-Grid             & 83.88\gaingray{0.03} & 85.79\gaingray{1.35} & 85.26\gaingray{1.20}   \\
+ Grad-Only                & 83.74\downgray{0.11} & 84.53\gaingray{0.09} & 83.87\downgray{0.19}   \\
+ Entropy-Only             & 83.80\downgray{0.05} & 84.31\downgray{0.13} & 84.00\downgray{0.06}   \\
\midrule
\rowcolor{tableblue!80}
\textbf{+ \method-\lite} 
& 83.97\gainblue{0.12} & 87.52\gainblue{3.08} & 85.56\gainblue{1.50}  \\
\rowcolor{tablepeach!70}
\textbf{+ \method-\feat}
& 83.89\gainpeach{0.04} & 87.13\gainpeach{2.69} & 84.99\gainpeach{0.93}  \\
\bottomrule
\end{tabular}
}
\label{tab:sampler}
\vspace{-8pt}
\end{wraptable}

\textbf{Comparison with Simple Non-random Samplers.}
We further examine whether the gains can be explained by simply replacing random cropping with a non-random heuristic.
As reported in Table~\ref{tab:sampler}, center-biased, uniform-grid, gradient-only, and entropy-only sampling yield limited or inconsistent improvements across datasets.
In contrast, \method provides more consistent gains, suggesting that effective sub-volume selection benefits from jointly considering anatomical plausibility, informativeness, and diversity rather than any single heuristic.
Detailed experimental settings and sampler definitions are provided in Appendix~\ref{app:simple_sampler}.

\subsection{Understanding: How Do the Three Principles Shape Where to Look? (RQ4)}
\label{sec:understanding}

\subsubsection{Anatomy-aware Preference}
\label{sec:understanding_anatomy}
\textbf{Dataset-level Center Spatial Distribution.}

\begin{wrapfigure}{r}{0.56\textwidth}
\centering
\vspace{-24pt}
\includegraphics[width=1.0\linewidth]{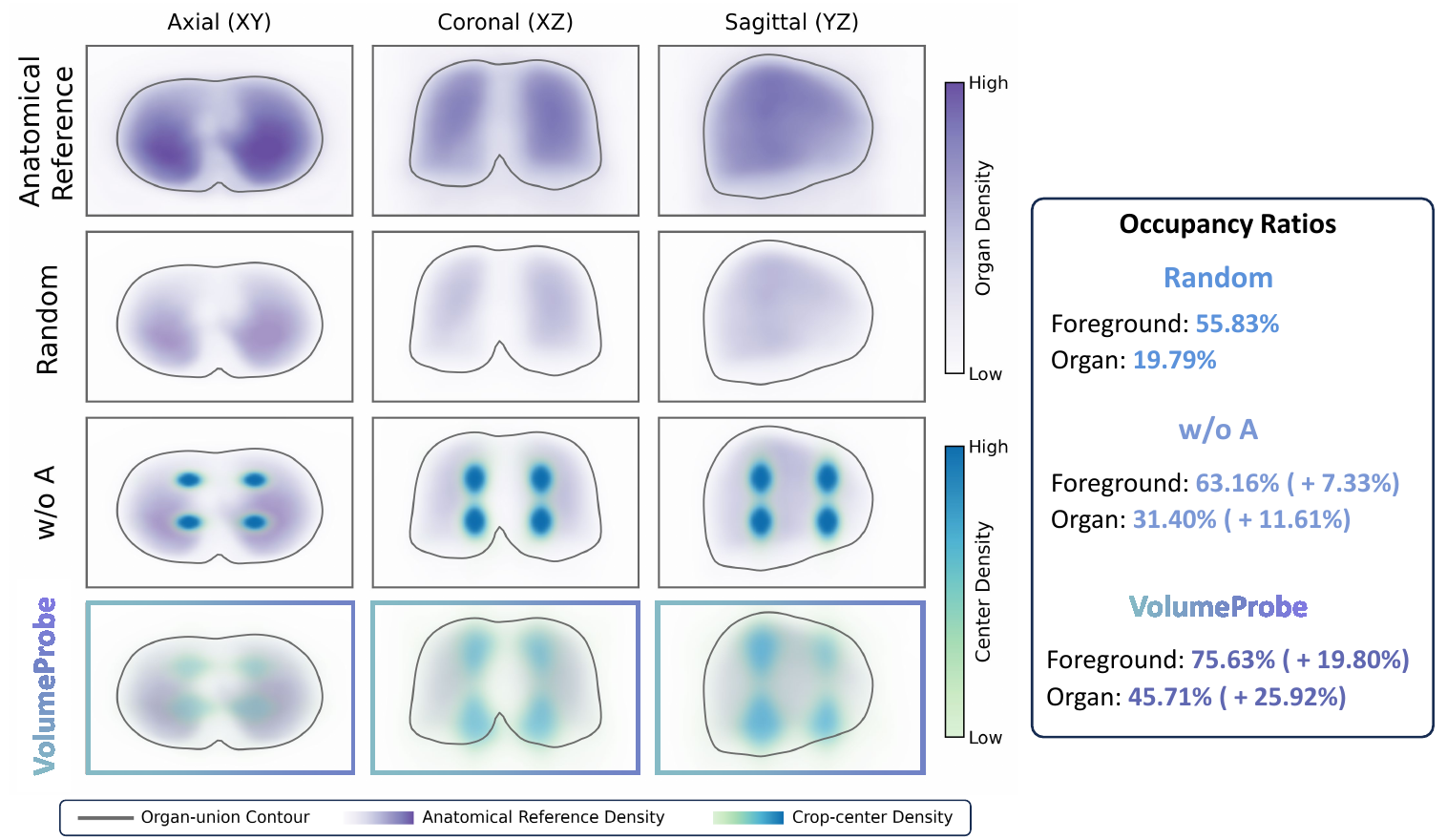}
\vspace{-12pt}
\caption{\textbf{Anatomy-aware analysis.} We aggregate organ masks to construct an anatomical reference density (purple), and overlay crop-center densities. \method better aligns crop centers with organ-dense regions and achieves higher foreground and organ occupancy.}
\label{fig:anatomy}
\vspace{-8pt}
\end{wrapfigure}

Fig.~\ref{fig:anatomy} compares dataset-level normalized crop center distributions. \textit{Random} spreads centers diffusely with limited overlap onto organ-dense regions, whereas \textit{w/o A} forms sharper hotspots driven mainly by non-anatomical cues. \method instead concentrates sub-crop centers within the organ-union contour and aligns more closely with the anatomical reference density, raising foreground occupancy from 55.83\% to 75.63\% and organ occupancy from 19.79\% to 45.71\% over \textit{Random}. The anatomy-aware term thus goes beyond a simple center bias, actively steering the probe bank toward \textbf{anatomically plausible and content-rich observations}.

\begin{figure*}
\centering
\includegraphics[width=1.00\linewidth]{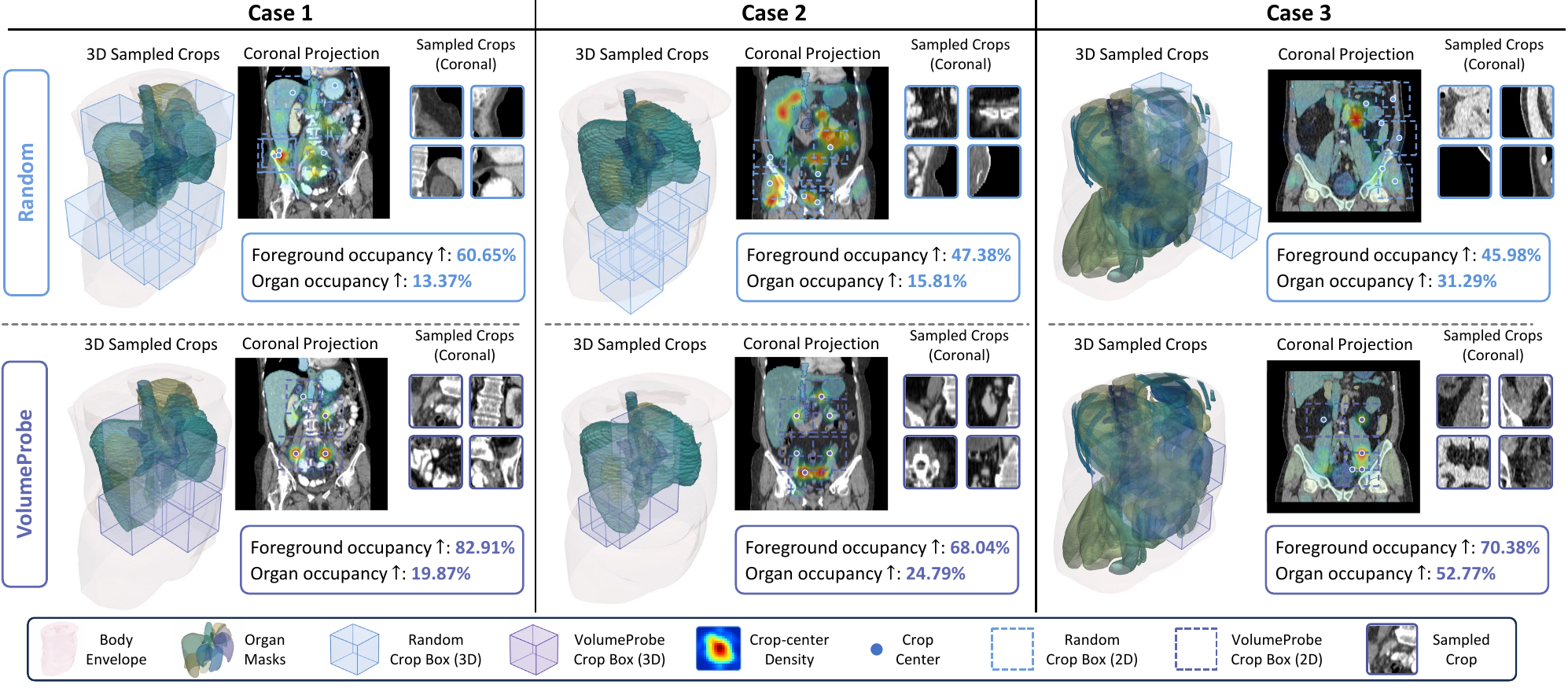}
\vspace{-12pt}
\caption{\textbf{Representative cases for anatomy-aware sampling.} We plot 3D crop boxes overlaid on organ masks, the coronal projection with sampled centers, 2D crop boxes, and example cropped views. Compared with \textit{Random}, \method places crops more consistently inside the body and around anatomically meaningful regions, raising foreground/organ occupancy ratios.}
\vspace{-6pt}
\label{fig:volume_vis}
\end{figure*} 

\textbf{Representative Case Studies for Anatomy-aware Sampling.}
Fig.~\ref{fig:volume_vis} provides case-level evidence across representative volumes from BTCV~\cite{landman2015miccai} and TotalSegmentator~\cite{wasserthal2023totalsegmentator}. \textit{Random} frequently places crops near body boundaries or partially outside the anatomical region, whereas \method concentrates them within the body interior, yielding higher foreground and organ occupancy ratios. 

\subsubsection{Informative Observations}
\label{sec:understanding_informative}

\begin{wrapfigure}{r}{0.50\textwidth}
  \centering
  \vspace{-14pt}
  \includegraphics[width=1.0\linewidth]{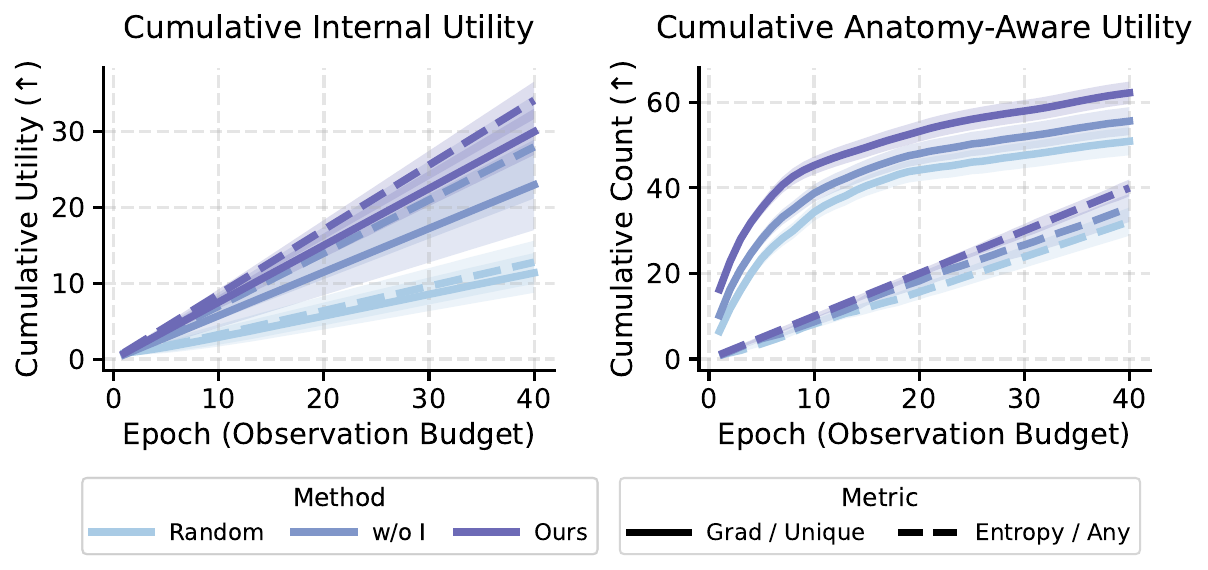}
  \vspace{-12pt}
  \caption{\textbf{Informativeness analysis}. \emph{Left}: cumulative internal utility, measured by gradient- and entropy-based metrics. \emph{Right}: cumulative anatomy-aware utility measured by the number of distinct organs observed and any-organ hits. }
  \label{fig:observation}
  \vspace{-8pt}
\end{wrapfigure}

We examine whether the informativeness term facilitates content-richer observations using both intrinsic image statistics and an external anatomical signal from TotalSegmentator~\cite{wasserthal2023totalsegmentator}. As shown in Fig.~\ref{fig:observation}, \method consistently accumulates higher gradient/entropy utility than \textit{Random} and \textit{w/o I}, while also observing more distinct organs and more any-organ hits despite not optimizing these anatomical labels. This highlights that \textbf{informativeness not only increases local structural content, but also improves the anatomical usefulness of the selected observations}, providing a stronger basis for the learning gains analyzed in Sec.~\ref{sec:learning}.

\subsubsection{Static Diversity and Redundancy}
\label{sec:understanding_diversity}

\begin{wrapfigure}{r}{0.54\textwidth}
  \centering
  \vspace{-22pt}
  \includegraphics[width=1.0\linewidth]{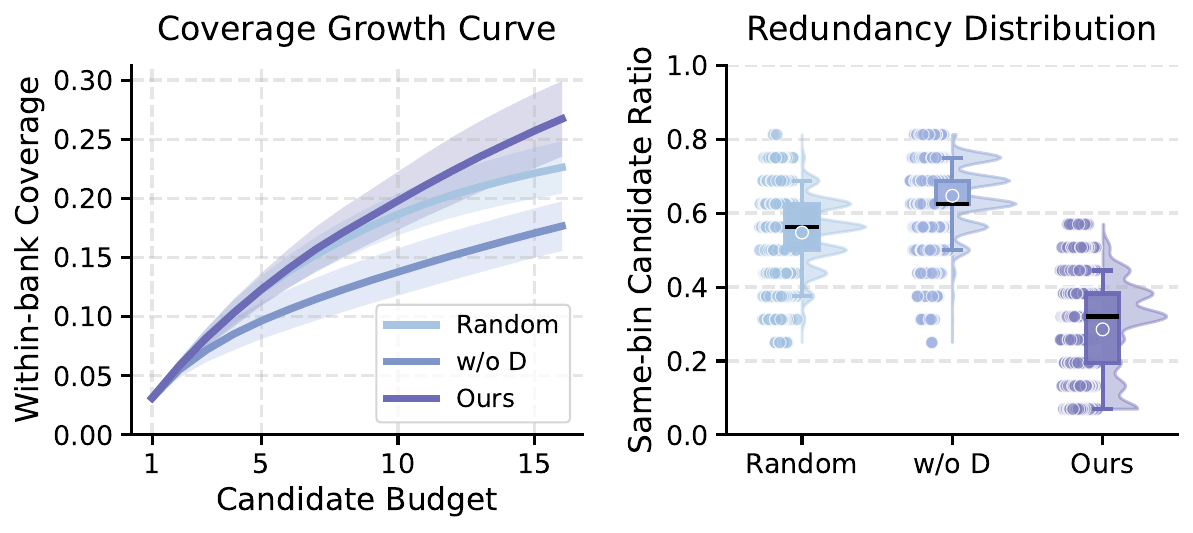}
  \vspace{-12pt}
  \caption{\textbf{Diversity analysis.} \textit{Left:} within-bank coverage as candidate budget increases. \method consistently achieves broader support. \textit{Right:} same-bin candidate ratio measuring local redundancy. Our approach yields the lowest redundancy. }
  \label{fig:diversity}
  \vspace{-8pt}
\end{wrapfigure}

We analyze whether the diversity term prevents the cached bank from concentrating around a few high-score regions. Fig.~\ref{fig:diversity} showcases that removing diversity (\textit{w/o D}) reduces support coverage and increases same-bin redundancy, even falling below \textit{Random} in coverage, which suggests that informative scoring alone tends to collapse candidates into locally salient areas. In contrast, \method maintains broader spatial support with lower redundancy, indicating that diversity is essential for \textbf{preventing bank collapse} rather than merely spreading candidates apart.




\subsection{Learning Relevance: From Better Crops to Better Representations (RQ5)}
\label{sec:learning}

\begin{figure*}[t]
\centering
\includegraphics[width=1.00\linewidth]{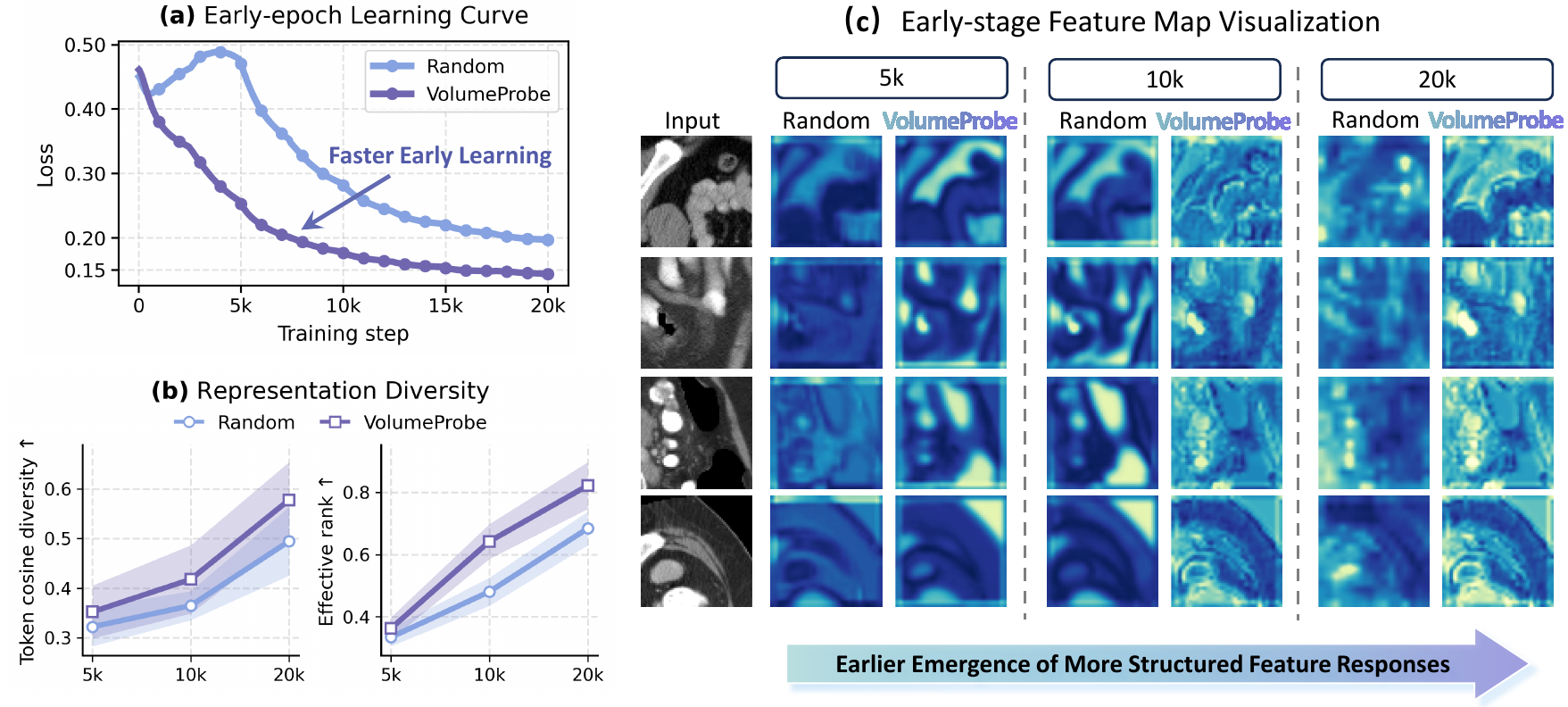}
\vspace{-12pt}
\caption{\textbf{Learning relevance of \method.}
\textbf{(a)} Early-stage learning curves, where \method shows faster and more stable loss reduction.
\textbf{(b)} Early representation diversity measured by token-level cosine diversity and effective rank.
\textbf{(c)} Feature-map visualization, where \method produces more structure-aligned representations.}
\vspace{-12pt}
\label{fig:learning}
\end{figure*}

Sec.~\ref{sec:understanding} has shown that \method reshapes \emph{where}  the model looks. We now explore whether these improved observations affect  \emph{how} the model learns. By focusing on early pretraining stages 
(up to 20k steps, i.e., 20\% of pretraining progress) of VoCo~\cite{Wu_2024_CVPR}, we isolate the effect of observation quality before convergence narrows inter-method differences, thus closing the loop with Sec.~\ref{sec:motivation}.

\begin{wrapfigure}{r}{0.52\textwidth}
  \centering
  \vspace{-6pt}
  \includegraphics[width=1.0\linewidth]{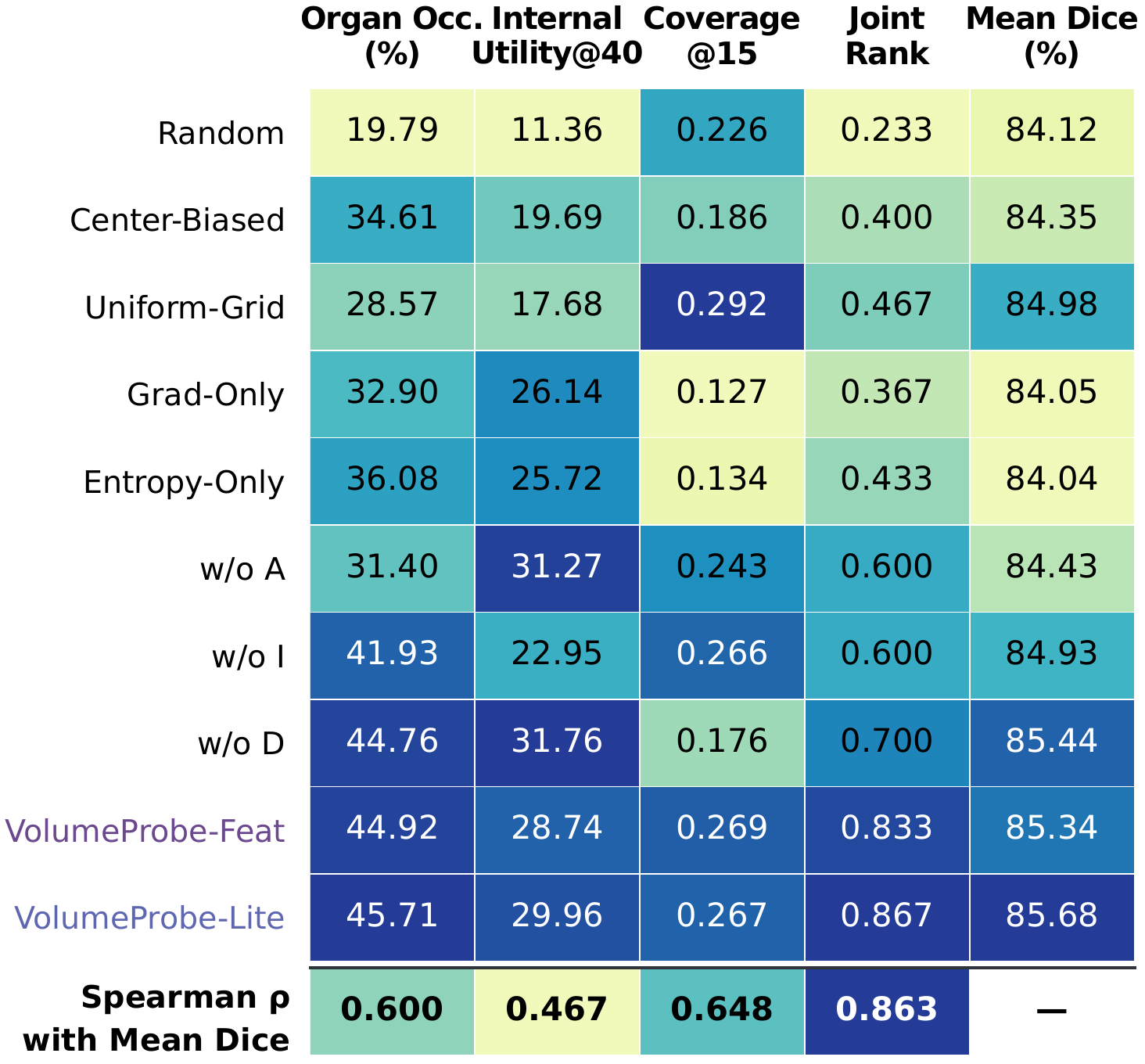}
  \vspace{-12pt}
    \caption{\textbf{Cross-strategy association between mechanistic diagnostics and downstream transfer.}}
  \label{fig:cross_strategy_heatmap}
  \vspace{-16pt}
\end{wrapfigure}

\textbf{Faster and More Stable Optimization.}
As shown in Fig.~\ref{fig:learning}(a), \method enters a stable loss descent earlier than \textit{Random} and maintains a lower pretraining loss throughout early training. It reaches the loss of \textit{Random}@20\% pretraining progress at only 7.9\% progress, indicating that the cached probe bank provides more efficient optimization signals under the same crop budget. This advantage persists under matched extended pretraining while retaining better compute efficiency (Appendix~\ref{app:extended_pretrain}).

\textbf{Less Collapsed Representations.}
Beyond optimization, Fig.~\ref{fig:learning}(b) indicates that \method also improves early representation geometry, yielding higher token-level cosine diversity and effective rank across checkpoints. The former measures pairwise non-redundancy among token features, while the latter captures the global spread of the representation covariance. Their consistent gains suggest that \method induces less redundant spatial features and a broader embedding spectrum.

\textbf{Structure-aligned Responses.}
We further probe the trained encoders in Fig.~\ref{fig:learning}(c). As shown, \method-trained models produce less diffuse and more anatomy-aligned feature responses than \textit{Random}. Additional quantitative metrics in Appendix~\ref{app:additional_learning_relevance} further confirm this trend across mask-available volumes. Together, these results verify that \textbf{better probing behavior not only accelerates early optimization, but also promotes less collapsed and more structure-aware representations}.

\textbf{Linking Mechanistic Diagnostics to Downstream Transfer.}
We finally examine whether the mechanistic diagnostics above track downstream utility across ten samplers from Tables~\ref{tab:ablation} and~\ref{tab:sampler}, without additional pretraining or fine-tuning.
We consider the following diagnostics:

\begin{itemize}
    \setlength{\itemsep}{0pt}
    \setlength{\parskip}{0pt}
    \setlength{\parsep}{0pt}
    
    \item \textbf{Organ Occupancy} (Sec.~\ref{sec:understanding_anatomy}) measures the fraction of sampled crop voxels overlapping annotated organ masks, reflecting anatomical validity.
    
    \item \textbf{Internal Utility@40} (Sec.~\ref{sec:understanding_informative}) measures cumulative local structural information over 40 observations, defined as
    $\mathrm{Utility@40}=\sum_{t=1}^{40}(\widetilde{\mathrm{Grad}}_t+\widetilde{\mathrm{Ent}}_t)/2$,
    where gradient magnitude and entropy are normalized within each volume.
    
    \item \textbf{Coverage@15} (Sec.~\ref{sec:understanding_diversity}) measures the spatial support covered by 15 selected centers, normalized to $[0,1]$, reflecting sampling coverage.
    
    \item \textbf{Joint Rank} equally combines the percentile ranks of the three diagnostics:
    $\mathrm{JointRank}_i=(R_{\mathrm{Organ},i}+R_{\mathrm{Utility},i}+R_{\mathrm{Coverage},i})/3$.
    
    \item \textbf{Mean Dice} averages the matched 100\%-label Dice on BTCV, AMOS, and WORD, serving as the downstream transfer measure.
\end{itemize}

Fig.~\ref{fig:cross_strategy_heatmap} summarizes the cross-strategy comparison across the ten sampling methods, with original values annotated in each cell.
Individual diagnostics show only moderate correlations with Mean Dice ($\rho=0.600$, $0.467$, and $0.648$); for example, high coverage or utility alone does not consistently translate into better transfer. In contrast, Joint Rank correlates more strongly with downstream performance ($\rho=0.863$, exact $p=0.0022$), remaining stable under leave-one-strategy-out analysis ($\rho=0.812$--$0.912$).
\textbf{Downstream transfer is therefore most consistently associated with a balanced combination of anatomical validity, local informativeness, and spatial coverage rather than any single diagnostic alone.}

\section{Related Work}

\textbf{3D Medical Image SSL.}
Self-supervised learning has emerged as a major paradigm for 3D medical image analysis, largely to alleviate the scarcity of annotations~\cite{Wu_2024_CVPR, Rui_2025_CVPR, 11122336, zhao2026towards, luo2025invcoss}. Existing methods are predominantly built on contrastive learning~\cite{Wu_2024_CVPR, Pan_2025_ICCV, He_2025_ICCV}, masked image modeling~\cite{TANG2026103770, fu2026medgmae, Chen_2023_WACV, 10977020, 11408915}, or their hybrids, refining representation learning through multi-view augmentation, contextual priors or reconstruction~\cite{gao2026learning, liu2026meduet, NI2026104027}, which have advanced label-efficient transfer across downstream tasks~\cite{10617802, 11274411, dai2025orochi, WANG2026103992}. Yet most efforts concentrate on learning from sampled sub-volumes while taking the sampling process itself as given. The implicit assumption that uniformly random sub-volumes constitute adequate observations remains largely unquestioned, despite well-known heterogeneity of medical volumes in tissue composition, anatomical structure, and foreground density~\cite{LEI2025103370, lee2025hu, huang2026vasomim, zhao2025large, 10980990}, leaving the observation policy underlying sub-crop pretraining underexplored.

\textbf{Spatial Observation Allocation in SSL.}
Prior work on data selection in SSL mainly studies \textit{which} training samples deserve more attention, through curriculum learning, self-paced learning, or importance sampling~\cite{10.1145/1553374.1553380,NEURIPS2024_ff8d608f}, operating at the dataset level and leaving intra-sample spatial selection unexplored. In 2D SSL, non-uniform observation has been explored via object-aware or foreground/saliency-sensitive view construction~\cite{NEURIPS2021_65d2ea03, li2024genview, Li_2022_CVPR}, attention-based region selection~\cite{Chen_2023_CVPR, ZHU2025111324, kakogeorgiou2022hide}, video keyframe selection~\cite{Tang_2025_CVPR,Wang_2025_CVPR,guo2025logicinframes, Zhang_2025_ICCV, Li_2026_WACV, Cao_2025_ICCV}, and optimized masking in masked image modeling~\cite{Wang_2023_CVPR, Bandara_2023_CVPR, xiang2026aemim, ishikawa2024data, hondru2025masked, 10960509, NEURIPS2022_5c186016}. In 3D medical imaging, anatomical priors, spatial redundancy, anisotropic spacing, and sub-crop pretraining together make \emph{where-to-look} within a volume a distinct problem~\cite{Wu_2024_CVPR,TANG2026103770,tang2024hyspark,lee2025hu}. While some medical SSL methods encode spatial structure in the pretext task~\cite{Wu_2024_CVPR,11274411}, the observation mechanism itself remains random cropping.

\section{Conclusion}
\label{sec:conclusion}
In this paper, we shed light on sub-volume sampling, treating it not merely as a preprocessing step but as a structured \emph{where-to-look} problem in a fixed-crop setting. We propose \method, a simple yet effective module that replaces unconstrained random cropping with a probe bank guided by anatomy-awareness, informativeness, and diversity. Experiments across multiple SSL frameworks show that \method redirects sub-volume sampling toward higher-quality observations, yielding consistent gains in transfer performance and more efficient learning. We hope this work helps illuminate sub-volume observation as an underexplored design dimension in 3D medical SSL.

\noindent\textbf{Limitations.}
\method builds a static probe bank offline and therefore cannot adapt its probing to the model's evolving learning dynamics. Future work will explore training-aware probe selection and more adaptive spatial priors beyond the non-canonical fields of view evaluated here.





\bibliographystyle{abbrv} 
\bibliography{neurips_2026}


\clearpage
\appendix
\section*{Appendix}

\section{Datasets}
\label{app:data}
\subsection{Pretraining Datasets}

For each SSL backbone, we use the same pretraining dataset as in its original work and follow the corresponding official implementation to ensure a fair and consistent comparison.

\textbf{VoCo}~\cite{Wu_2024_CVPR}. 
Following the default VoCo setup, we pretrain VoCo on VoCo-10k~\cite{Wu_2024_CVPR}, a large-scale CT collection comprising BTCV~\cite{landman2015miccai}, TCIA Covid19~\cite{clark2013cancer}, LUNA16~\cite{SETIO20171}, STOIC 2021~\cite{doi:10.1148/radiol.2021210384}, TotalSegmentator~\cite{wasserthal2023totalsegmentator}, FLARE23~\cite{ma2024automatic}, and HNSCC~\cite{grossberg2018imaging}. In total, VoCo-10k contains approximately 9.9k CT volumes. We use the released open-source data splits from prior work~\cite{Wu_2024_CVPR, 11004165, 10977020}. A detailed summary of the pretraining datasets is provided in Table~\ref{tab:pretrain_data}.

\textbf{S2DC}~\cite{Pan_2025_ICCV}. 
Following the original S2DC setting, we also use VoCo-10k~\cite{Wu_2024_CVPR} as the pretraining dataset, resulting in approximately 9.9k CT volumes.

\textbf{MedGMAE}~\cite{fu2026medgmae}. 
For MedGMAE, we follow its original pretraining setup and use AbdomenAtlasMini 1.0~\cite{LI2024103285}, which contains about 5.2k CT volumes, as summarized in Table~\ref{tab:pretrain_data}.

\begin{table*}[h]
\centering
\caption{Overview of pretraining datasets.}
\setlength{\tabcolsep}{1.5mm}
\renewcommand{\arraystretch}{1.1}
\resizebox{0.70\linewidth}{!}{
\begin{tabular}{lccc}
\toprule
\textbf{Dataset} & \textbf{Modality} & \textbf{Region of Interest} & \textbf{\#Volumes} \\
\midrule
BTCV~\cite{landman2015miccai} & CT & Abdomen                     & 24   \\
TCIA Covid19~\cite{clark2013cancer} & CT & Chest                       & 722  \\
LUNA16~\cite{SETIO20171} & CT & Chest                       & 843  \\
STOIC 2021~\citep{doi:10.1148/radiol.2021210384} & CT & Chest                       & 2,000 \\
FLARE23~\cite{ma2024automatic} & CT & Abdomen                     & 4,000 \\
HNSCC~\cite{grossberg2018imaging} & CT & Head/Neck                   & 1,071 \\
TotalSegmentator~\cite{wasserthal2023totalsegmentator} & CT & \makecell[l]{Head/Neck/Chest/Leg/ \\Abdomen/Pelvis/Feet} & 1,203 \\
AbdomenAtlasMini 1.0~\cite{LI2024103285} & CT & Abdomen                     & 5,195 \\
\bottomrule
\end{tabular}
}
\label{tab:pretrain_data}
\end{table*}

\subsection{Downstream Datasets}

Details of the downstream datasets and splits are provided in Table~\ref{tab:downstream_data}. We evaluate the pretrained representations on three downstream task families. For segmentation, we use BTCV~\cite{landman2015miccai}, AMOS~\cite{NEURIPS2022_ee604e1b}, WORD~\cite{LUO2022102642}, and Atlas-MRI~\cite{data8050079}. For classification, we use CC-CCII~\cite{zhang2020clinically}. For registration, we use IXI~\cite{KIM2021102036} and OASIS~\cite{6795067}. Following prior work~\citep{Wu_2024_CVPR, 11274411, fu2026medgmae, TANG2026103770}, we adopt the established data splits for each benchmark whenever available to ensure comparable evaluation.

\begin{table*}[t]
\centering
\caption{Overview of downstream datasets.}
\setlength{\tabcolsep}{2.0mm}
\resizebox{0.80\linewidth}{!}{
\begin{tabular}{lllccccc}
\toprule
\multirow{2}{*}{\textbf{Task}} 
& \multirow{2}{*}{\textbf{Dataset}} 
& \multirow{2}{*}{\textbf{Modality}} 
& \multicolumn{4}{c}{\textbf{\#Train}} 
& \multirow{2}{*}{\textbf{\#Valid}} \\
\cmidrule(lr){4-7}
& & & Full & 1\% & 10\% & 100\% & \\
\midrule
\multirow{4}{*}{Segmentation}
& BTCV~\cite{landman2015miccai} & CT & 24  & 1 & 2  & 24  & 6   \\
& AMOS~\cite{NEURIPS2022_ee604e1b} & CT & 200 & 2 & 20 & 200 & 100 \\
& WORD~\citep{LUO2022102642} & CT & 100 & 1 & 10 & 100 & 20  \\
& Atlas-MRI~\cite{data8050079} & MRI & 45 & 1 & 5 & 45 & 15 \\
\midrule
Classification
& CC-CCII~\citep{zhang2020clinically} & CT & 2,514 & -- & -- & -- & 1,664 \\
\midrule
\multirow{2}{*}{Registration}
& IXI~\cite{KIM2021102036} & MRI & 403 & -- & -- & -- & 115 \\
& OASIS~\cite{6795067} & MRI & 394 & -- & -- & -- & 19 \\
\bottomrule
\end{tabular}
}
\label{tab:downstream_data}
\end{table*}

\section{Implementation Details}
\label{app:imple}

\subsection{Pretraining Setup}
Table~\ref{tab:pretrain_settings} presents the baseline-specific preprocessing and pretraining settings. We strictly follow the official setup of VoCo~\cite{Wu_2024_CVPR}, S2DC~\cite{Pan_2025_ICCV}, and MedGMAE~\cite{fu2026medgmae}, including the preprocessing pipeline, augmentation strategy, optimizer, learning-rate schedule, and pretraining length. This ensures that the comparison isolates the effect of the sub-volume sampling policy rather than changes in backbone-specific training strategies. The key differences are their sub-volume configurations, namely, VoCo uses one $64^3$ crop, S2DC uses one $96^3$ crop, and MedGMAE uses three $96^3$ crops per volume. For MedGMAE, we independently sample three crop centers either from the cached probe bank or via standard random cropping at each iteration, without imposing an additional diversity constraint. This configuration keeps MedGMAE's original three-crop training scheme unchanged, while replacing only the source of crop centers. 

During pretraining, we sample 70\% of crop centers from the cached probe bank and the remaining 30\% via standard random cropping. For bank-sampled centers, we apply a small local jitter independently along each spatial axis. Specifically, the center is shifted by a uniformly sampled integer offset within \(\pm \mathrm{round}(\gamma r_a)\), where \(r_a\) is the sub-volume size on axis \(a\) and \(\gamma=0.05\) by default, and the jittered center is clipped to the valid image range. All methods are pretrained for 100k steps with AdamW and a warmup cosine schedule. \method only replaces random sampling with cached probe-bank sampling, leaving all other settings unchanged.

\subsection{Downstream Settings}
For segmentation and classification, we follow the same preprocessing and fine-tuning protocols as prior work~\cite{Wu_2024_CVPR, 11274411}, as summarized in Table~\ref{tab:downstream_settings}. Specifically, VoCo~\cite{Wu_2024_CVPR} and S2DC~\cite{Pan_2025_ICCV} use Swin-UNETR~\cite{hatamizadeh2021swin} as the downstream backbone, while MedGMAE~\cite{fu2026medgmae} uses UNETR~\cite{Hatamizadeh_2022_WACV}. Following previous work~\cite{TANG2026103770, fu2026medgmae}, we fine-tune segmentation models for 5000, 2000, and 1000 epochs under 1\%, 10\%, and 100\% label regimes, respectively. For registration, we use TransMorph~\cite{CHEN2022102615} as the downstream baseline, with preprocessing and training protocols kept consistent with VoCo~\cite{Wu_2024_CVPR} for fair comparison.

\begin{table}[t]
\centering
\caption{Overview of preprocessing and pretraining settings used for pretraining.}
\setlength{\tabcolsep}{3.0mm}
\renewcommand{\arraystretch}{1.1}
\resizebox{1.0\linewidth}{!}{
\begin{tabular}{lccc}
\toprule
\textbf{Setting} & \textbf{VoCo}~\cite{Wu_2024_CVPR} & \textbf{S2DC}~\cite{Pan_2025_ICCV} & \textbf{MedGMAE}~\cite{fu2026medgmae} \\
\midrule
\rowcolor{tablegray!80}
\multicolumn{4}{c}{\textit{Preprocessing Settings}} \\
Spacing           & $1.5\times1.5\times1.5$ & $1.5\times1.5\times1.5$ & $1.5\times1.5\times1.5$ \\
Intensity         & $[-175, 250]$            & $[-175, 250]$            & $[-175, 250]$ \\
Norm              & $[0, 1]$                 & $[0, 1]$                 & $[0, 1]$ \\
Sub-volume size   & $64\times64\times64$     & $96\times96\times96$     & $96\times96\times96$ \\
Sub-volume number & 1                         & 1                         & 3 \\
Augmentation      & Random Rotate and Flip    & Random Rotate and Flip    & Random Rotate, Flip, Scale, Shift \\
\midrule
\rowcolor{tablegray!80}
\multicolumn{4}{c}{\textit{Pretraining Settings}} \\
Pretraining steps  & 100k & 100k & 100k \\
Optimizer          & AdamW & AdamW & AdamW \\
Weight decay       & 1e-2 & 1e-1 & 0 \\
Optimizer momentum & $\beta_1, \beta_2 = 0.9, 0.999$ & $\beta_1, \beta_2 = 0.9, 0.999$ & $\beta_1, \beta_2 = 0.9, 0.95$ \\
Optimizer LR       & 1e-3 & 4e-4 & 1e-5 \\
Batch size         & 4 & 4 & 8 \\
LR schedule        & Warmup cosine & Warmup cosine & Warmup cosine \\
Warmup steps       & 2k & 0.5k & 2k \\
\bottomrule
\end{tabular}
}
\label{tab:pretrain_settings}
\end{table}

\begin{table}[t]
\centering
\caption{Overview of preprocessing and fine-tuning settings for downstream evaluation.}
\setlength{\tabcolsep}{3.0mm}
\renewcommand{\arraystretch}{1.1}
\resizebox{0.50\linewidth}{!}{
\begin{tabular}{lc}
\toprule
\rowcolor{tablegray!80}
\multicolumn{2}{c}{\textit{Preprocessing Settings}} \\
Spacing          & $1.5\times1.5\times1.5$ \\
Intensity        & $[-175, 250]$ \\
Norm             & $[0, 1]$ \\
Sub-volume size  & $96\times96\times96$ \\
Sub-volume number   & 4 \\
Augmentation     & \makecell[c]{Random Rotate, Flip,\\Scale, Shift} \\
\midrule
\rowcolor{tablegray!80}
\multicolumn{2}{c}{\textit{Fine-tuning Settings}} \\
Optimizer            & AdamW \\
Weight decay         & 1e-5 \\
Optimizer momentum   & $\beta_{1}, \beta_{2}=0.9, 0.999$ \\
Optimizer LR         & 3e-4 \\
Batch size           & $1\times4 = 4$ \\
Inference            & Sliding Window Inference \\
Inference overlap    & 0.75 \\
Sw batch size        & 4 \\
\bottomrule
\end{tabular}
}
\label{tab:downstream_settings}
\end{table}

\subsection{Heterogeneity of Evaluated SSL Frameworks}
Although all evaluated medical SSL frameworks~\cite{Wu_2024_CVPR, Pan_2025_ICCV, fu2026medgmae} leverage sub-volume cropping, this is exactly the interface targeted by our \method. Notably, the three frameworks cover different SSL objectives, downstream architectures, and crop configurations. Concretely, VoCo~\cite{Wu_2024_CVPR} and S2DC~\cite{Pan_2025_ICCV} are contrastive-style methods using Swin-UNETR~\cite{hatamizadeh2021swin}, whereas MedGMAE~\cite{fu2026medgmae} follows masked modeling and uses UNETR~\cite{Hatamizadeh_2022_WACV}. They also differ in crop size and number, with VoCo using one $64^3$ crop, S2DC using one $96^3$ crop, and MedGMAE using three $96^3$ crops per volume. These differences highlight that \method is evaluated across distinct objectives, architectures, and single- versus multi-crop interfaces, supporting its plug-and-play applicability to representative 3D medical SSL pipelines.

\subsection{Frozen Feature Encoder for \method-\feat}
For \method-\feat, we use 3DINO-ViT~\cite{xu2025generalizable} as a frozen feature encoder only for offline probe-bank construction. 3DINO-ViT is pretrained on a large-scale multimodal 3D medical imaging corpus of about 100k 3D scans, covering CT, MRI, and PET scans across diverse anatomical regions, rather than on natural images. We regard \method-\feat as an external-feature variant and use it mainly to study whether richer semantic features further improve probe selection. It is worth noting that \method-\lite does not use any external encoder and already shows consistent gains, so our main conclusion does not rely on 3DINO-ViT pretraining data. Moreover, rather than introducing new cues, we highlight observation policy as an underexplored aspect and show that simple lightweight cues of \method-\lite already suffice under this \emph{where-to-look} formulation.

\begin{figure*}[t]
\centering
\includegraphics[width=1.00\linewidth]{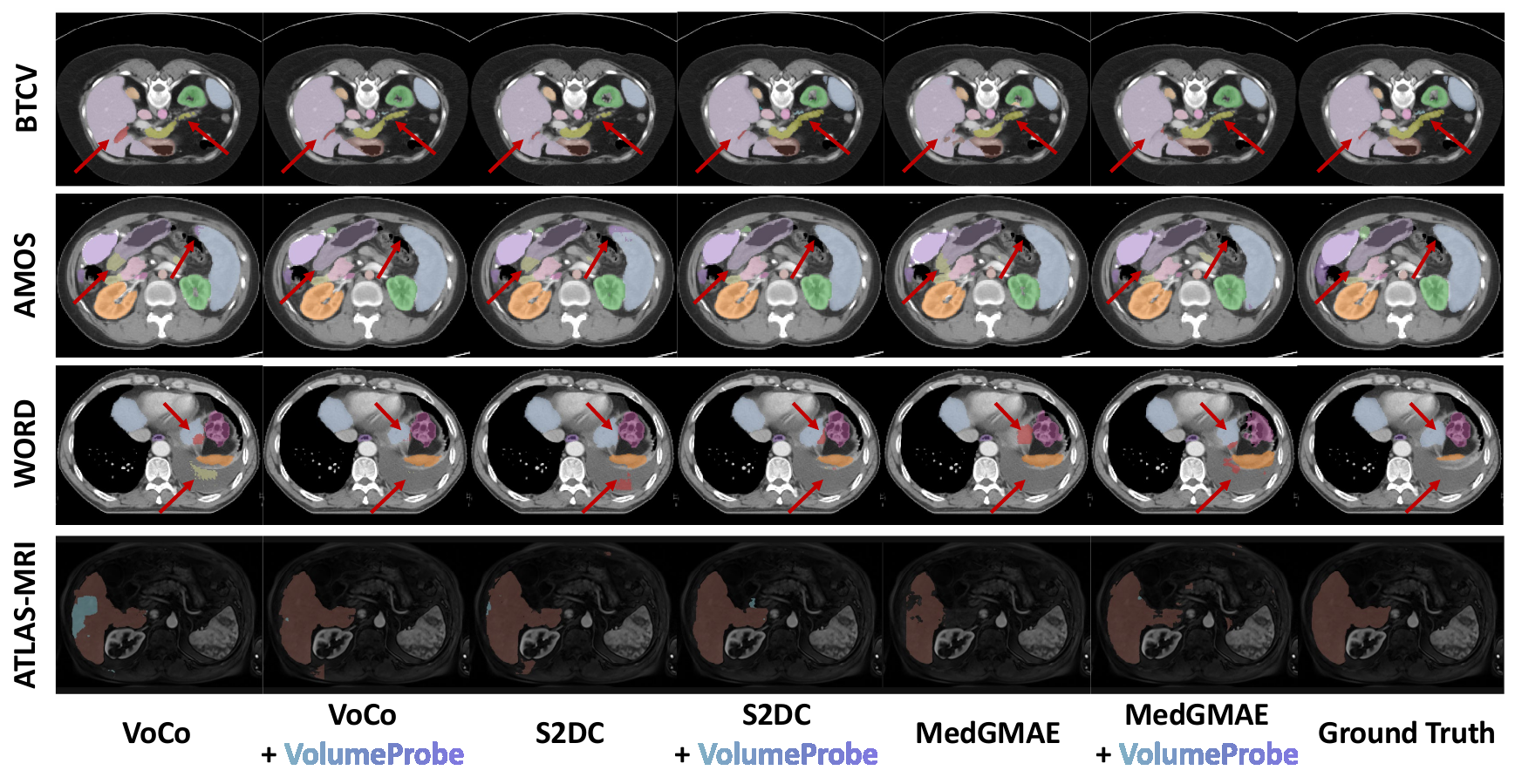}
\caption{Qualitative visualization of segmentation results under the 10\% label setting.}
\label{fig:visual_seg}
\vspace{-8pt}
\end{figure*} 

\section{More Experimental Results}
\subsection{Multi-run Robustness Analysis}
\label{app:multi-run_robustness}
We further evaluate whether the main downstream gains are sensitive to random seeds. For the 100\% label setting and registration, we keep the official train/test splits fixed and repeat downstream training with three different optimization seeds. For the 1\% and 10\% label settings, we use three labeled-subset seeds, where each seed defines the same labeled subset for the random-cropping baseline, \method-\lite and \method-\feat, together with the corresponding downstream optimization seed. This paired protocol isolates the effect of the sampling policy while avoiding changes to the official evaluation split.

As reported in Tables~\ref{tab:std_seg},~\ref{tab:std_cls}, and~\ref{tab:std_reg}, \method consistently improves the random-cropping baselines across segmentation, classification, and registration under multi-run evaluation. The variance is larger in low-label segmentation and classification, while the 100\% label and registration settings remain relatively stable. The observed gains remain consistent across runs, indicating that \method is robust to seed-level variation across downstream tasks and SSL backbones.

\begin{table*}[t]
\centering
\setlength{\tabcolsep}{1.0mm}
\caption{\textbf{Multi-run robustness on segmentation tasks.}
Dice Score (\%) is reported as mean$_{\pm \mathrm{std}}$ over three independent runs.}
\vspace{-5pt}
\renewcommand\arraystretch{1.00}
\resizebox{1.0\linewidth}{!}{
\begin{tabular}{lccc|ccc|ccc|ccc}
\toprule
\multirow{3}{*}{\textbf{Method}}
& \multicolumn{12}{c}{\textbf{Dice Score (\%) $\uparrow$}} \\
\cmidrule(lr){2-13}
& \multicolumn{3}{c|}{\textbf{BTCV}} 
& \multicolumn{3}{c|}{\textbf{AMOS}} 
& \multicolumn{3}{c|}{\textbf{WORD}} 
& \multicolumn{3}{c}{\textbf{Atlas-MRI}} \\
\cmidrule(lr){2-4}\cmidrule(lr){5-7}\cmidrule(lr){8-10}\cmidrule(lr){11-13}
& 1\% & 10\% & 100\%
& 1\% & 10\% & 100\%
& 1\% & 10\% & 100\%
& 1\% & 10\% & 100\% \\
\midrule
VoCo~\cite{Wu_2024_CVPR}
  & 63.33\stdgray{1.5} & 77.85\stdgray{0.7} & 83.85\stdgray{0.2}
  & 55.81\stdgray{2.3} & 73.34\stdgray{0.6} & 84.44\stdgray{0.1}
  & 64.24\stdgray{1.2} & 77.28\stdgray{0.4} & 84.06\stdgray{0.2}
  & 34.85\stdgray{1.9} & 47.56\stdgray{0.9} & 58.83\stdgray{0.3} \\
\rowcolor{tableblue!80}
\textbf{+ \method-\lite} 
  & 67.21\stdblue{1.3} & \underline{79.39}\stdblue{0.5} & 83.97\stdblue{0.1}
  & 58.64\stdblue{1.5} & 74.95\stdblue{0.6} & \textbf{87.52}\stdblue{0.1}
  & 68.93\stdblue{1.6} & \textbf{80.12}\stdblue{0.5} & \textbf{85.56}\stdblue{0.2}
  & 38.68\stdblue{2.2} & 49.27\stdblue{1.0} & 59.98\stdblue{0.2} \\
\rowcolor{tablepeach!70}
\textbf{+ \method-\feat}
  & 68.07\stdpeach{1.5} & \textbf{79.51}\stdpeach{0.9} & 83.89\stdpeach{0.1}
  & 57.86\stdpeach{1.9} & 74.70\stdpeach{0.4} & 87.13\stdpeach{0.1}
  & 68.42\stdpeach{1.7} & 79.74\stdpeach{0.4} & 84.99\stdpeach{0.2}
  & 37.93\stdpeach{2.4} & 49.06\stdpeach{1.2} & 59.42\stdpeach{0.2} \\
\midrule
S2DC~\cite{Pan_2025_ICCV}
  & 66.77\stdgray{1.8} & 77.93\stdgray{0.8} & 84.09\stdgray{0.1}
  & 55.19\stdgray{1.6} & 74.28\stdgray{0.9} & 84.68\stdgray{0.2}
  & 69.14\stdgray{2.0} & 78.96\stdgray{0.6} & 84.46\stdgray{0.1}
  & 33.79\stdgray{2.1} & 46.18\stdgray{0.8} & 58.53\stdgray{0.2} \\
\rowcolor{tableblue!80}
\textbf{+ \method-\lite} 
  & \textbf{69.47}\stdblue{1.5} & 78.74\stdblue{0.9} & \textbf{84.43}\stdblue{0.1}
  & 58.83\stdblue{2.1} & 75.31\stdblue{0.7} & \underline{87.17}\stdblue{0.1}
  & 69.73\stdblue{1.5} & \underline{80.04}\stdblue{0.3} & \underline{85.22}\stdblue{0.1}
  & 37.16\stdblue{2.7} & 48.08\stdblue{0.9} & 59.44\stdblue{0.3} \\
\rowcolor{tablepeach!70}
\textbf{+ \method-\feat}
  & \underline{68.76}\stdpeach{1.3} & 78.92\stdpeach{1.0} & \underline{84.19}\stdpeach{0.1}
  & 59.47\stdpeach{1.4} & 75.17\stdpeach{0.3} & 86.87\stdpeach{0.1}
  & 70.28\stdpeach{1.9} & 79.82\stdpeach{0.5} & 85.43\stdpeach{0.2}
  & 38.19\stdpeach{2.2} & 48.83\stdpeach{1.0} & 59.81\stdpeach{0.2} \\
\midrule
MedGMAE~\cite{fu2026medgmae}
  & 66.19\stdgray{2.0} & 77.11\stdgray{1.1} & 83.22\stdgray{0.2}
  & 58.79\stdgray{2.4} & 75.65\stdgray{0.8} & 84.90\stdgray{0.1}
  & 68.09\stdgray{1.7} & 77.73\stdgray{0.6} & 84.19\stdgray{0.1}
  & 36.24\stdgray{2.9} & 48.16\stdgray{1.2} & 59.42\stdgray{0.3} \\
\rowcolor{tableblue!80}
\textbf{+ \method-\lite} 
  & 66.85\stdblue{1.7} & 78.44\stdblue{1.0} & 83.48\stdblue{0.1}
  & \textbf{60.26}\stdblue{1.9} & \textbf{77.14}\stdblue{0.9} & 85.62\stdblue{0.1}
  & \underline{70.67}\stdblue{1.9} & 79.42\stdblue{0.7} & 84.36\stdblue{0.1}
  & \textbf{39.71}\stdblue{2.5} & \textbf{50.23}\stdblue{1.2} & \underline{60.19}\stdblue{0.3} \\
\rowcolor{tablepeach!70}
\textbf{+ \method-\feat}
  & 66.57\stdpeach{2.1} & 79.09\stdpeach{0.7} & 83.29\stdpeach{0.1}
  & \underline{59.57}\stdpeach{1.5} & \underline{76.80}\stdpeach{0.6} & 85.93\stdpeach{0.2}
  & \textbf{71.15}\stdpeach{1.4} & 79.08\stdpeach{0.9} & 84.25\stdpeach{0.1}
  & \underline{39.48}\stdpeach{2.2} & \underline{49.83}\stdpeach{1.0} & \textbf{60.40}\stdpeach{0.2} \\
\bottomrule
\end{tabular}
}
\label{tab:std_seg}
\end{table*}

\begin{table*}[t]
\centering
\begin{minipage}[t]{0.48\textwidth} 
\centering 
\caption{\textbf{Multi-run robustness on classification tasks.}} 
\setlength{\tabcolsep}{2.5mm} 
\renewcommand\arraystretch{1.00} 
\resizebox{0.80\linewidth}{!}{ 
\begin{tabular}{lc} 
\toprule 
\multirow{2}{*}{\textbf{Method}} & \textbf{Accuracy (\%) $\uparrow$} \\ 
\cmidrule(lr){2-2} & \textbf{CC-CCII} \\ 
\midrule 
VoCo~\cite{Wu_2024_CVPR} & 93.45\stdgray{0.4} \\ 
\rowcolor{tableblue!80} 
\textbf{+ \method-\lite} & 93.98\stdblue{0.9} \\
\rowcolor{tablepeach!70} 
\textbf{+ \method-\feat} & 94.25\stdpeach{0.5} \\
\midrule 
S2DC~\cite{Pan_2025_ICCV} & 94.19\stdgray{1.1} \\
\rowcolor{tableblue!80} 
\textbf{+ \method-\lite} & \textbf{94.87}\stdblue{0.8} \\
\rowcolor{tablepeach!70} 
\textbf{+ \method-\feat} & \underline{94.34}\stdpeach{1.2} \\
\midrule 
MedGMAE~\cite{fu2026medgmae} & 93.08\stdgray{0.6} \\
\rowcolor{tableblue!80} 
\textbf{+ \method-\lite} & 93.58\stdblue{0.5} \\
\rowcolor{tablepeach!70} 
\textbf{+ \method-\feat} & 93.81\stdpeach{0.6} \\
\bottomrule 
\end{tabular} 
} 
\label{tab:std_cls} 
\end{minipage}
\hfill
\begin{minipage}[t]{0.48\textwidth}
    \centering
    \caption{\textbf{Multi-run robustness on registration tasks.}}
    \label{tab:std_reg}
    \setlength{\tabcolsep}{2.5mm}
    \renewcommand\arraystretch{1.00}
    \resizebox{0.91\linewidth}{!}{
    \begin{tabular}{lcc}
    \toprule
    \multirow{2}{*}{\textbf{Method}} & \multicolumn{2}{c}{\textbf{Dice Score (\%) $\uparrow$}} \\
    \cmidrule(lr){2-3}
     & \textbf{IXI} & \textbf{OASIS} \\
    \midrule
    VoCo~\cite{Wu_2024_CVPR} & 73.66\stdgray{0.4} & 84.38\stdgray{0.3} \\
    \rowcolor{tableblue!80}
    \textbf{+ \method-\lite} & 74.39\stdblue{0.3} & 85.55\stdblue{0.5} \\
    \rowcolor{tablepeach!70}
    \textbf{+ \method-\feat} & 74.17\stdpeach{0.4} & 85.19\stdpeach{0.3} \\
    \midrule
    S2DC~\cite{Pan_2025_ICCV} & 72.93\stdgray{0.2} & 82.97\stdgray{0.4} \\
    \rowcolor{tableblue!80}
    \textbf{+ \method-\lite} & 73.70\stdblue{0.4} & 84.09\stdblue{0.6} \\
    \rowcolor{tablepeach!70}
    \textbf{+ \method-\feat} & 73.98\stdpeach{0.3} & 83.86\stdpeach{0.6} \\
    \midrule
    MedGMAE~\cite{fu2026medgmae} & 73.69\stdgray{0.5} & 85.71\stdgray{0.3} \\
    \rowcolor{tableblue!80}
    \textbf{+ \method-\lite} & \textbf{74.88}\stdblue{0.2} & \underline{85.96}\stdblue{0.4} \\
    \rowcolor{tablepeach!70}
    \textbf{+ \method-\feat} & 74.28\stdpeach{0.4} & \textbf{86.17}\stdpeach{0.2} \\
    \bottomrule
    \end{tabular}
    }
\end{minipage}
\end{table*}

\subsection{Subject-level Variability and Paired Significance}

To complement the run-level robustness analysis, we further conduct a
subject-level analysis over the same three matched downstream runs.
For each test subject, we first average its Dice scores across the three runs, and then compute the mean and standard deviation across subjects. Statistical significance is evaluated using a two-sided paired permutation test between \method-\lite and its corresponding random-cropping baseline. This protocol evaluates whether the observed improvements remain consistent at the subject level rather than only across optimization runs.
As shown in Table~\ref{tab:subject_significance}, \method reaches $p\leq0.05$ in 33 of the 36 paired
comparisons, while all mean improvements remain positive.
The results therefore indicate that the gains are consistent at the
subject level across the matched three-run aggregate.

\begin{table*}[t]
\centering
\setlength{\tabcolsep}{1.0mm}
\caption{\textbf{Subject-level variability and paired significance on segmentation tasks.}
Dice Score (\%) is reported with subject-level SD in subscript and significance in superscript. Superscripts denote two-sided paired permutation tests against the corresponding random-cropping baseline:
n.s., $p>0.05$; *, $p\leq0.05$; **, $p\leq0.01$;
***, $p\leq0.001$; ****, $p\leq0.0001$. For BTCV ($N=6$), the minimum exact two-sided $p$-value is $0.03125$, so significance is capped at *.}
\renewcommand\arraystretch{1.05}
\resizebox{1.0\linewidth}{!}{
\begin{tabular}{lccc|ccc|ccc|ccc}
\toprule
\multirow{3}{*}{\textbf{Method}}
& \multicolumn{12}{c}{\textbf{Dice Score (\%) $\uparrow$}} \\
\cmidrule(lr){2-13}
& \multicolumn{3}{c|}{\textbf{BTCV}}
& \multicolumn{3}{c|}{\textbf{AMOS}}
& \multicolumn{3}{c|}{\textbf{WORD}}
& \multicolumn{3}{c}{\textbf{Atlas-MRI}} \\
\cmidrule(lr){2-4}
\cmidrule(lr){5-7}
\cmidrule(lr){8-10}
\cmidrule(lr){11-13}
& 1\% & 10\% & 100\%
& 1\% & 10\% & 100\%
& 1\% & 10\% & 100\%
& 1\% & 10\% & 100\% \\
\midrule

VoCo~\cite{Wu_2024_CVPR}
& \statcellgray{63.33}{}{5.9}
& \statcellgray{77.85}{}{7.2}
& \statcellgray{83.85}{}{1.8}
& \statcellgray{55.81}{}{3.9}
& \statcellgray{73.34}{}{4.5}
& \statcellgray{84.44}{}{1.6}
& \statcellgray{64.24}{}{3.2}
& \statcellgray{77.28}{}{2.9}
& \statcellgray{84.06}{}{1.2}
& \statcellgray{34.85}{}{5.5}
& \statcellgray{47.56}{}{4.3}
& \statcellgray{58.83}{}{1.2} \\

\rowcolor{tableblue!80} 
\textbf{+ \method}
& \statcellblue{67.21}{*}{6.8}
& \statcellblue{79.39}{*}{8.6}
& \statcellblue{83.97}{n.s.}{1.3}
& \statcellblue{58.64}{***}{4.0}
& \statcellblue{74.95}{****}{4.2}
& \statcellblue{87.52}{****}{1.4}
& \statcellblue{68.93}{****}{3.7}
& \statcellblue{80.12}{****}{3.5}
& \statcellblue{85.56}{****}{1.5}
& \statcellblue{38.68}{**}{4.2}
& \statcellblue{49.27}{n.s.}{3.5}
& \statcellblue{59.98}{**}{1.0} \\

\midrule

S2DC~\cite{Pan_2025_ICCV}
& \statcellgray{66.77}{}{8.2}
& \statcellgray{77.93}{}{5.7}
& \statcellgray{84.09}{}{2.3}
& \statcellgray{55.19}{}{4.2}
& \statcellgray{74.28}{}{3.9}
& \statcellgray{84.68}{}{1.5}
& \statcellgray{69.14}{}{4.0}
& \statcellgray{78.96}{}{4.3}
& \statcellgray{84.46}{}{1.3}
& \statcellgray{33.79}{}{4.6}
& \statcellgray{46.18}{}{5.7}
& \statcellgray{58.53}{}{1.0} \\

\rowcolor{tableblue!80} 
\textbf{+ \method}
& \statcellblue{69.47}{*}{7.4}
& \statcellblue{78.74}{*}{5.6}
& \statcellblue{84.43}{*}{1.5}
& \statcellblue{58.83}{****}{4.0}
& \statcellblue{75.31}{****}{4.1}
& \statcellblue{87.17}{****}{1.9}
& \statcellblue{69.73}{***}{3.5}
& \statcellblue{80.04}{***}{3.9}
& \statcellblue{85.22}{****}{1.0}
& \statcellblue{37.16}{**}{5.0}
& \statcellblue{48.08}{**}{3.8}
& \statcellblue{59.44}{**}{0.9} \\

\midrule

MedGMAE~\cite{fu2026medgmae}
& \statcellgray{66.19}{}{5.9}
& \statcellgray{77.11}{}{6.9}
& \statcellgray{83.22}{}{1.9}
& \statcellgray{58.79}{}{4.2}
& \statcellgray{75.65}{}{5.0}
& \statcellgray{84.90}{}{2.1}
& \statcellgray{68.09}{}{3.4}
& \statcellgray{77.73}{}{2.9}
& \statcellgray{84.19}{}{1.2}
& \statcellgray{36.24}{}{4.3}
& \statcellgray{48.16}{}{3.9}
& \statcellgray{59.42}{}{1.2} \\

\rowcolor{tableblue!80} 
\textbf{+ \method}
& \statcellblue{66.85}{*}{6.3}
& \statcellblue{78.44}{n.s.}{7.2}
& \statcellblue{83.48}{*}{2.4}
& \statcellblue{60.26}{****}{4.3}
& \statcellblue{77.14}{****}{4.4}
& \statcellblue{85.62}{***}{2.2}
& \statcellblue{70.67}{***}{3.9}
& \statcellblue{79.42}{***}{3.5}
& \statcellblue{84.36}{****}{1.5}
& \statcellblue{39.71}{**}{4.0}
& \statcellblue{50.23}{**}{4.2}
& \statcellblue{60.19}{***}{1.4} \\

\bottomrule
\end{tabular}
}
\label{tab:subject_significance}
\end{table*}

\subsection{Visualization Results}
\label{app:vis}
Fig.~\ref{fig:visual_seg} provides qualitative segmentation results under the 10\% label setting across four downstream datasets. Compared with the corresponding SSL baselines, adding \method produces more complete and anatomically aligned predictions. These visual results further support that improved sub-volume sampling leads to more transferable representations across different SSL backbones.

\subsection{Evaluation under nnU-Net Pipelines}
\label{app:nnunet}

Following~\cite{11274411}, we use the official nnU-Net pipeline~\cite{isensee2021nnu} and perform dataset fingerprinting, planning, and preprocessing separately for AMOS~\cite{NEURIPS2022_ee604e1b} and WORD~\cite{LUO2022102642}. Within each dataset, all variants share the same auto-configured 3D full-resolution plan, data split, and training protocol, while differing only in whether the pretrained weights are obtained using Random or \method sampling. Table~\ref{tab:nnunet_eval} confirms that \method consistently improves both VoCo~\cite{Wu_2024_CVPR} and S2DC~\cite{Pan_2025_ICCV} across the two datasets and label regimes, demonstrating that its benefits persist under a strong, systematically configured segmentation framework.

\begin{table*}[t]
\centering
\setlength{\tabcolsep}{2.0mm}
\caption{\textbf{Evaluation under nnU-Net pipelines.}
Subject-level Dice Score (\%) is reported as mean $\pm$ SD.
All variants within each dataset share the same auto-configured nnU-Net
3D full-resolution plan and training protocol.}
\renewcommand\arraystretch{1.0}
\resizebox{0.65\linewidth}{!}{
\begin{tabular}{lcc|cc}
\toprule
\multirow{2}{*}{\textbf{Method}}
& \multicolumn{4}{c}{\textbf{Dice Score (\%) $\uparrow$}} \\
\cmidrule(lr){2-5}
& \multicolumn{2}{c|}{\textbf{AMOS}}
& \multicolumn{2}{c}{\textbf{WORD}} \\
\cmidrule(lr){2-3}
\cmidrule(lr){4-5}
& 1\% & 100\%
& 1\% & 100\% \\
\midrule

nnU-Net (from scratch)
& 29.53\stdgray{4.7}
& 84.89\stdgray{1.9}
& 49.82\stdgray{3.3}
& 84.66\stdgray{1.5} \\

\midrule
VoCo (nnU-Net)~\cite{Wu_2024_CVPR}
& 52.96\stdgray{3.4}
& 85.22\stdgray{2.0}
& 65.47\stdgray{3.9}
& 84.89\stdgray{1.2} \\

\rowcolor{tableblue!80}
\textbf{+ \method}
& \textbf{55.43}\stdblue{3.5}
& \textbf{86.90}\stdblue{1.8}
& \textbf{67.53}\stdblue{4.2}
& \textbf{85.37}\stdblue{1.5} \\

\midrule
S2DC (nnU-Net)~\cite{Pan_2025_ICCV}
& 49.86\stdgray{2.8}
& 85.38\stdgray{2.2}
& 67.21\stdgray{3.1}
& 84.83\stdgray{1.4} \\

\rowcolor{tableblue!80}
\textbf{+ \method}
& \textbf{54.05}\stdblue{3.2}
& \textbf{86.82}\stdblue{2.0}
& \textbf{70.34}\stdblue{4.0}
& \textbf{85.46}\stdblue{1.1} \\

\bottomrule
\end{tabular}
}
\label{tab:nnunet_eval}
\end{table*}


\begin{table}[t]
\centering
\caption{\textbf{Occupancy ratios of sampled sub-volumes.} Foreground occupancy measures the average foreground voxel ratio within sampled crops, while organ occupancy measures the average organ-mask voxel ratio on the mask-available subset.}
\setlength{\tabcolsep}{2.0mm}
\renewcommand{\arraystretch}{1.1}
\resizebox{0.75\linewidth}{!}{
\begin{tabular}{lcc}
\toprule
\textbf{Method} & \textbf{Foreground Occupancy (\%) $\uparrow$} & \textbf{Organ Occupancy (\%) $\uparrow$} \\
\midrule
Random & 55.83 & 19.79 \\
\midrule
Center-Biased & 69.12\gaingray{13.29} & 34.61\gaingray{14.82} \\
Uniform-Grid & 62.74\gaingray{6.91} & 28.57\gaingray{8.78} \\
Grad-Only & 71.34\gaingray{15.51} & 32.90\gaingray{13.11} \\
Entropy-Only & 68.76\gaingray{12.93} & 36.08\gaingray{16.29} \\
\midrule
\rowcolor{tableblue!80}
\method-\lite & 75.63\gainblue{19.80} & 45.71\gainblue{25.92} \\
\rowcolor{tablepeach!70}
\method-\feat & 77.19\gainpeach{21.36} & 44.92\gainpeach{25.13} \\
\bottomrule
\end{tabular}
}
\label{tab:occupancy_ratio}
\end{table}

\subsection{Experimental Details for Simple Non-random Samplers}
\label{app:simple_sampler}

We provide additional experimental details for the simple non-random samplers compared in Table~\ref{tab:sampler}.
All samplers require no labels, learnable parameters, or online model feedback. For a fair comparison, each method follows the same pretraining setup as \method. For every preprocessed volume, we construct an offline bank of \(K\) crop centers and follow the same mixed sampling policy as \method, with 70\% bank-based sampling and 30\% standard random cropping during pretraining. The crop size, number of crops, SSL backbone, augmentation, optimizer, learning-rate schedule, and total pretraining steps are kept unchanged across all sampling strategies.

We consider four simple non-random policies:
\begin{itemize}
    \item \textbf{Center-Biased.} Selects candidates closest to the center of the foreground-cropped volume, isolating the effect of a simple centrality prior.
    \item \textbf{Uniform-Grid.} Distributes crop centers approximately uniformly over the valid crop-center space, isolating spatial coverage.
    \item \textbf{Grad-Only.} Ranks candidates by the mean local gradient magnitude, capturing local structural variation.
    \item \textbf{Entropy-Only.} Ranks candidates by local intensity entropy, capturing local intensity heterogeneity.
\end{itemize}
Together, these policies isolate simple spatial and image-statistical cues while keeping the remaining pretraining pipeline unchanged.

\begin{table*}[t]
\centering

\begin{minipage}[t]{0.46\textwidth}
    \centering
    \caption{\textbf{Additional ablation of principles on classification and registration tasks.}}
    \label{tab:ablation_cls_reg}
    \setlength{\tabcolsep}{1.5mm}
    \renewcommand\arraystretch{1.0}
    \resizebox{1.00\linewidth}{!}{
    \begin{tabular}{lcc}
    \toprule
    \textbf{Method}
    & \textbf{CC-CCII Acc. (\%) $\uparrow$}
    & \textbf{IXI Dice (\%) $\uparrow$} \\
    \midrule
    VoCo~\cite{Wu_2024_CVPR}
        & 93.45
        & 73.66 \\
    \rowcolor{tableblue!80}
    \textbf{+ \method}
        & \textbf{93.98}\gainblue{0.53}
        & \textbf{74.39}\gainblue{0.73} \\
    w/o $A$
        & 93.53\gaingray{0.08}
        & 74.05\gaingray{0.39} \\
    w/o $I$
        & 93.61\gaingray{0.16}
        & 73.59\downgray{0.07} \\
    w/o $D$
        & 93.78\gaingray{0.33}
        & 74.12\gaingray{0.46} \\
    \bottomrule
    \end{tabular}
    }
\end{minipage}
\hfill
\begin{minipage}[t]{0.50\textwidth}
    \centering
    \caption{\textbf{Relative gain retained after removing each criterion.}}
    \label{tab:ablation_gain_retained}
    \setlength{\tabcolsep}{1.5mm}
    \renewcommand\arraystretch{1.10}
    \resizebox{1.00\linewidth}{!}{
    \begin{tabular}{lccc}
    \toprule
    \textbf{Removed Criterion}
    & \textbf{Seg. (\%)}
    & \textbf{Cls. (\%)}
    & \textbf{Reg. (\%)} \\
    \midrule
    w/o $A$ & 20.2 & 15.1 & 53.4 \\
    w/o $I$ & 51.9 & 30.2 & -9.6 \\
    w/o $D$ & 84.5 & 62.3 & 63.0 \\
    \bottomrule
    \end{tabular}
    }
\end{minipage}

\end{table*}

\begin{table}[t]
\centering
\caption{Fine-grained ablation of principle terms for VoCo~\cite{Wu_2024_CVPR}. We remove each term from the anatomy-aware or informativeness score while keeping the remaining pipeline unchanged.}
\setlength{\tabcolsep}{2.0mm}
\resizebox{0.90\linewidth}{!}{
\begin{tabular}{lccccccc}
\toprule
\multirow{2}{*}{\textbf{Variant}} 
& \multicolumn{2}{c}{\textbf{Anatomy-aware}} 
& \multicolumn{2}{c}{\textbf{Informative}} 
& \multicolumn{3}{c}{\textbf{Dice Score (\%) $\uparrow$}} \\
\cmidrule(lr){2-3} \cmidrule(lr){4-5} \cmidrule(lr){6-8}
& \textbf{LocPrior} & \textbf{Margin} 
& \textbf{Grad} & \textbf{Entropy} 
& \textbf{BTCV} & \textbf{AMOS} & \textbf{WORD} \\
\midrule
VoCo~\cite{Wu_2024_CVPR} 
& -- & -- & -- & -- 
& 83.85 & 84.44 & 84.06  \\
\rowcolor{tableblue!80}
+ \method 
& \checkmark & \checkmark & \checkmark & \checkmark 
& 83.97\gainblue{0.12} & 87.52\gainblue{3.08} & 85.56\gainblue{1.50}  \\
\midrule
w/o LocPrior 
& $\times$ & \checkmark & \checkmark & \checkmark 
& 83.90\gaingray{0.05} & 85.60\gaingray{1.16} & 84.21\gaingray{0.15} \\
w/o Margin 
& \checkmark & $\times$ & \checkmark & \checkmark 
& 83.87\gaingray{0.02} & 86.73\gaingray{2.29} & 85.40\gaingray{1.34} \\
w/o Grad 
& \checkmark & \checkmark & $\times$ & \checkmark 
& 83.91\gaingray{0.06} & 87.13\gaingray{2.69} & 85.36\gaingray{1.30} \\
w/o Entropy 
& \checkmark & \checkmark & \checkmark & $\times$ 
& 83.95\gaingray{0.10} & 86.97\gaingray{2.53} & 85.47\gaingray{1.41} \\
\bottomrule
\end{tabular}
}
\label{tab:principle_terms}
\end{table}

\begin{table*}[t]
\centering
\caption{Ablation study of coarse-bin partition strategy for VoCo~\cite{Wu_2024_CVPR}.}
\setlength{\tabcolsep}{2.5mm}
\renewcommand\arraystretch{1.0}
\resizebox{0.55\linewidth}{!}{
\begin{tabular}{cccc}
\toprule
\multirow{2}{*}{\textbf{Coarse Bins Partition}} & \multicolumn{3}{c}{\textbf{Dice Score (\%) $\uparrow$}} \\
\cmidrule(lr){2-4}
& \textbf{BTCV} & \textbf{AMOS} & \textbf{WORD} \\
\midrule
$(G_x, G_y, G_z)$ \\
\midrule
$(3, 3, 1)$              & 81.90 & 84.46 & 83.80  \\
$(3, 3, 2)$              & 82.57 & 84.51 & 83.71  \\
$(4, 4, 1)$              & 83.40 & 86.27 & 85.09  \\
\rowcolor{tableblue!80}
$(4, 4, 2)$              & 83.97 & 87.52 & 85.56   \\
$(5, 5, 1)$              & 83.26 & 86.75 & 84.92  \\
$(5, 5, 2)$              & 83.89 & 87.04 & 85.37  \\
\bottomrule
\end{tabular}
}
\label{tab:parition}
\end{table*}

\begin{table*}[t]
\centering
\begin{minipage}[t]{0.48\textwidth}
    \centering
    \caption{Ablation study of $M$ for VoCo~\cite{Wu_2024_CVPR}.}
    \setlength{\tabcolsep}{2.5mm}
    \renewcommand\arraystretch{1.1}
    \resizebox{0.85\linewidth}{!}{
    \begin{tabular}{c|ccc}
    \toprule
    \multirow{2}{*}{$\mathbf{M}$} & \multicolumn{3}{c}{\textbf{Dice Score (\%) $\uparrow$}} \\
    \cmidrule(lr){2-4}
    & \textbf{BTCV} & \textbf{AMOS} & \textbf{WORD} \\
    \midrule
    $8$               & 82.16 & 84.83 & 83.71  \\
    $16$              & 82.79 & 86.12 & 84.29  \\
    \rowcolor{tableblue!80}
    $24$              & 83.97 & 87.52 & 85.56  \\
    $32$              & 83.71 & 87.60 & 85.11  \\
    \bottomrule
    \end{tabular}
    }
    \label{tab:ablation_m}
\end{minipage}
\hfill
\begin{minipage}[t]{0.46\textwidth}
    \centering
    \caption{Ablation study of $K$ for VoCo~\cite{Wu_2024_CVPR}.}
    \setlength{\tabcolsep}{2.5mm}
    \renewcommand\arraystretch{1.0}
    \resizebox{0.85\linewidth}{!}{
    \begin{tabular}{c|ccc}
    \toprule
    \multirow{2}{*}{$\mathbf{K}$} & \multicolumn{3}{c}{\textbf{Dice Score (\%) $\uparrow$}} \\
    \cmidrule(lr){2-4}
    & \textbf{BTCV} & \textbf{AMOS} & \textbf{WORD} \\
    \midrule
    $10$              & 81.89 & 83.76 & 83.81  \\
    $20$              & 82.70 & 85.39 & 84.61  \\
    $30$              & 83.61 & 87.10 & 85.24  \\
    \rowcolor{tableblue!80}
    $40$              & 83.97 & 87.52 & 85.56   \\
    $50$              & 83.88 & 87.29 & 85.62  \\
    \bottomrule
    \end{tabular}
    }
    \label{tab:ablation_k}
\end{minipage}
\end{table*}

\subsection{Occupancy Analysis across Sampling Strategies}
\label{app:sampler_occupancy}

To complement the downstream comparison in Table~\ref{tab:sampler}, we further examine how different sampling strategies allocate the sub-volume crop budget across anatomically valid regions.
We report the foreground occupancy ratio and organ occupancy ratio on VoCo-10k~\cite{Wu_2024_CVPR, 11274411} in Table~\ref{tab:occupancy_ratio}.
Foreground occupancy measures the fraction of crop voxels within the body foreground, while organ occupancy measures the fraction overlapping annotated organs on the mask-available subset, including BTCV~\cite{landman2015miccai}, LUNA16~\cite{SETIO20171}, FLARE23~\cite{ma2024automatic}, and TotalSegmentator~\cite{wasserthal2023totalsegmentator}.

\textit{Random} achieves only 55.83\% foreground occupancy and 19.79\% organ occupancy, while simple non-random policies improve these ratios to varying degrees. Nevertheless, they remain below \method-\lite, which reaches 75.63\% foreground occupancy and 45.71\% organ occupancy.
\method-\feat further increases foreground occupancy to 77.19\% while achieving 44.92\% organ occupancy. Taken together, simple spatial or statistical heuristics partially reduce observation waste, while the joint consideration of anatomical plausibility, informativeness, and diversity yields more anatomically valid and organ-rich observations.

\subsection{Ablation Study}
\label{app:ablation}
\textbf{Task-dependent contribution of the three principles.}
We further examine whether anatomy-awareness ($A$), informativeness ($I$), and diversity ($D$) contribute beyond segmentation. Following Sec.~\ref{sec:ablation}, we separately remove each criterion using VoCo~\cite{Wu_2024_CVPR} and evaluate on CC-CCII~\cite{zhang2020clinically} classification and IXI~\cite{KIM2021102036} registration, while keeping all other settings unchanged. As reported in Table~\ref{tab:ablation_cls_reg}, the complete objective achieves the best performance across both metrics. Removing informativeness causes the largest degradation on IXI, while removing anatomy-awareness leads to the largest reduction in classification performance.

To quantify task dependence, we further compute the fraction of the full \method gain retained after removing each criterion,
$(\mathrm{Ablation}-\mathrm{VoCo})/
(\mathrm{VolumeProbe}-\mathrm{VoCo})$.
For segmentation, we use the mean Dice gain over BTCV~\cite{landman2015miccai}, AMOS~\cite{NEURIPS2022_ee604e1b}, and WORD~\cite{LUO2022102642} under the 100\% label setting; for classification and registration, we use accuracy and Dice, respectively. Table~\ref{tab:ablation_gain_retained} indicates that
anatomy-awareness contributes more strongly to segmentation and classification, whereas informativeness is particularly important for registration. Diversity has a smaller but consistently beneficial contribution across the evaluated task families.

\textbf{Principle terms.}
Table~\ref{tab:principle_terms} provides a fine-grained ablation of the anatomy-aware and informativeness scores. Removing any individual term consistently weakens performance compared with the full \method, confirming that LocPrior, Margin, Grad, and Entropy each contribute to probe-bank quality. Among them, removing LocPrior causes the most pronounced degradation on AMOS~\cite{NEURIPS2022_ee604e1b} and WORD~\cite{LUO2022102642}, indicating that anatomical plausibility is especially important for selecting effective sub-volumes.

\textbf{Coarse-bin Partition}.
Table~\ref{tab:parition} studies the effect of the coarse partition resolution used in candidate generation. A moderately anisotropic grid, $(G_x,G_y,G_z)=(4,4,2)$, achieves the best performance across three downstream datasets, reflecting that it provides a favorable balance between spatial localization and candidate coverage. Coarser partitions overlook informative regions, while overly fine partitions introduce fragmented candidate pools and slightly degrade the probe-bank quality.

\textbf{Sensitivity to $M$ and $K$.}
Tables~\ref{tab:ablation_m} and~\ref{tab:ablation_k} verify the sensitivity to the number of retained coarse bins $M$ and probe-bank size $K$. For $M$, performance improves as more coarse bins are retained and peaks overall at $M=24$, while further increasing $M$ to 32 brings no consistent gain, confirming that overly many coarse candidates may introduce less informative regions. For $K$, enlarging the probe bank steadily improves performance up to $K=40$, after which the gains saturate and slightly fluctuate. We therefore use $M=24$ and $K=40$ as the default setting, which provides a good balance between candidate diversity and probe quality.

\textbf{Sensitivity to $\alpha$ and $\beta$.}
Tables~\ref{tab:ablation_alpha} and~\ref{tab:ablation_beta} study the sensitivity to the informativeness weight $\alpha$ and diversity weight $\beta$. Performance remains stable across $\alpha,\beta\in\{0.5,1.0,1.5\}$, with $\alpha=1.0$ and $\beta=1.0$ achieving the best overall results across BTCV, AMOS, and WORD. The stable performance confirms that \method is not overly sensitive to moderate changes in these weighting hyperparameters, and we use $\alpha=\beta=1.0$ by default.

\begin{table*}[t]
\centering
\begin{minipage}[t]{0.48\textwidth}
    \centering
    \caption{Ablation study of $\alpha$ for VoCo~\cite{Wu_2024_CVPR}.}
    \setlength{\tabcolsep}{2.5mm}
    \renewcommand\arraystretch{1.0}
    \resizebox{0.85\linewidth}{!}{
    \begin{tabular}{c|ccc}
    \toprule
    \multirow{2}{*}{$\mathbf{\alpha}$} & \multicolumn{3}{c}{\textbf{Dice Score (\%) $\uparrow$}} \\
    \cmidrule(lr){2-4}
    & \textbf{BTCV} & \textbf{AMOS} & \textbf{WORD} \\
    \midrule
    $0.5$              & 83.91 & 87.43 & 85.37  \\
    \rowcolor{tableblue!80}
    $1.0$              & 83.97 & 87.52 & 85.56   \\
    $1.5$              & 83.89 & 87.29 & 85.25  \\
    \bottomrule
    \end{tabular}
    }
    \label{tab:ablation_alpha}
\end{minipage}
\hfill
\begin{minipage}[t]{0.46\textwidth}
    \centering
    \caption{Ablation study of $\beta$ for VoCo~\cite{Wu_2024_CVPR}.}
    \setlength{\tabcolsep}{2.5mm}
    \renewcommand\arraystretch{1.0}
    \resizebox{0.85\linewidth}{!}{
    \begin{tabular}{c|ccc}
    \toprule
    \multirow{2}{*}{$\mathbf{\beta}$} & \multicolumn{3}{c}{\textbf{Dice Score (\%) $\uparrow$}} \\
    \cmidrule(lr){2-4}
    & \textbf{BTCV} & \textbf{AMOS} & \textbf{WORD} \\
    \midrule
    $0.5$              & 83.94 & 87.30 & 85.26  \\
    \rowcolor{tableblue!80}
    $1.0$              & 83.97 & 87.52 & 85.56   \\
    $1.5$              & 83.86 & 87.24 & 85.48  \\
    \bottomrule
    \end{tabular}
    }
    \label{tab:ablation_beta}
\end{minipage}
\end{table*}

\subsection{Convergence under Extended Pretraining}
\label{app:extended_pretrain}

\begin{figure*}[t]
    \centering
    \includegraphics[width=1.00\linewidth]{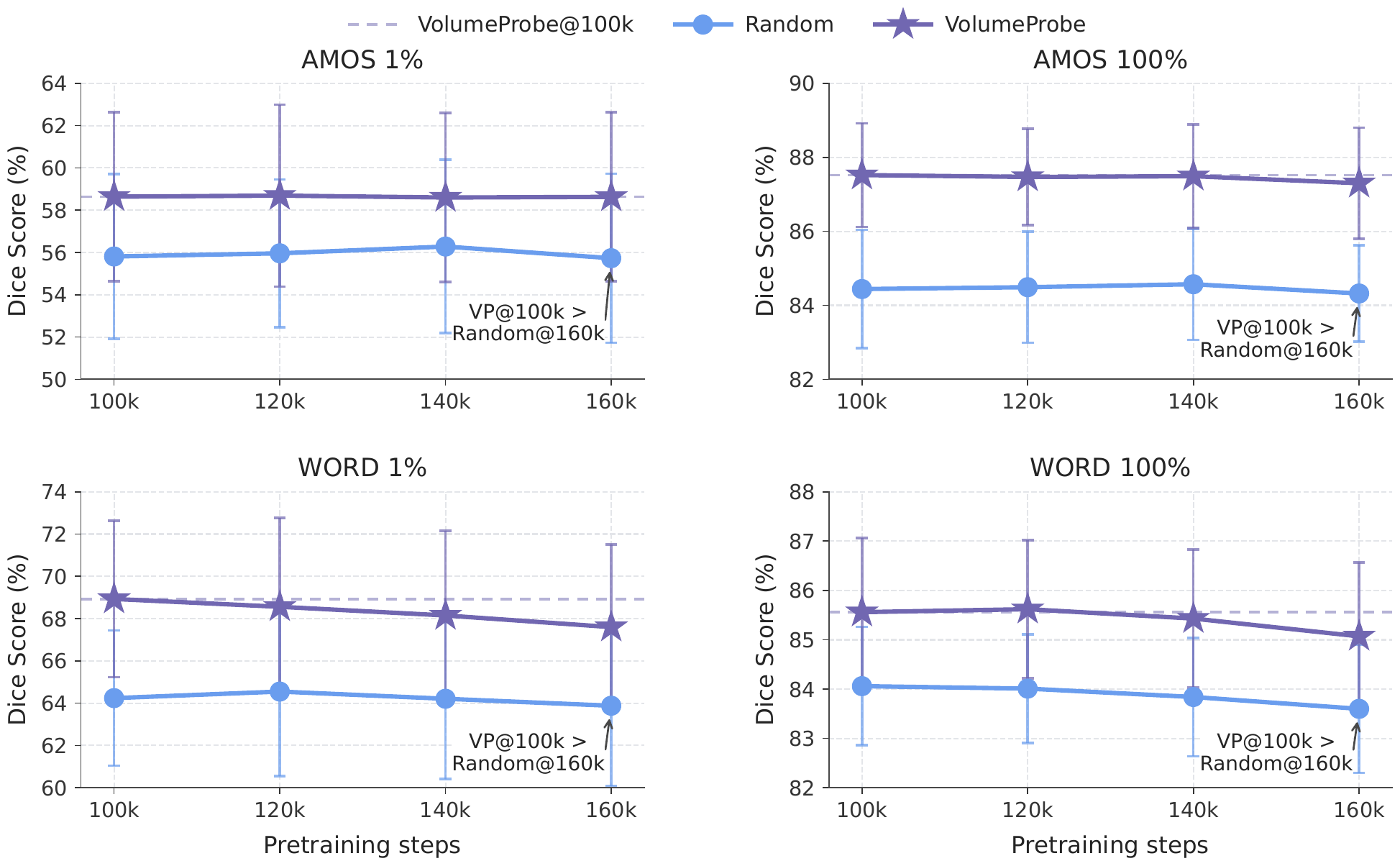}
    \caption{\textbf{Convergence under extended pretraining.}
    Downstream Dice of VoCo with \textit{Random} sampling and \method from 100k to 160k steps. \method consistently outperforms \textit{Random}; the gap narrows on WORD 1\% but remains stable elsewhere, and \method@100k still exceeds \textit{Random}@160k in all settings.}
    \label{fig:extended_pretraining_curves}
\end{figure*}

To examine whether the gains of \method mainly reflect faster convergence or persist under longer pretraining, we extend VoCo~\cite{Wu_2024_CVPR} pretraining from the original 100k steps to 160k steps under matched continuation schedules for both \textit{Random} sampling and \method. Starting from the 100k checkpoints, we resume both methods for an additional 60k steps with their optimizer states restored, using the same 2k-step warmup followed by cosine decay, while keeping all other settings unchanged.

Fig.~\ref{fig:extended_pretraining_curves} compares downstream Dice across the four matched checkpoints. \method consistently outperforms
\textit{Random} at every checkpoint in all four settings. The gap narrows on WORD 1\%, while remaining largely stable in the other three settings. Notably, \method@100k still outperforms \textit{Random}@160k in all four settings, indicating that longer \textit{Random} pretraining does not close the gap within the tested range.

The efficiency conclusion remains after accounting for total compute.
Including the one-time bank construction and the 4.4\% recurring overhead (Table~\ref{tab:efficiency_analysis}), \method@100k requires 284.9 GPU hours, compared with approximately 323.2 and 430.9 GPU hours for \textit{Random}@120k and \textit{Random}@160k, respectively. It therefore achieves better performance with 11.8\% and 33.9\% less total compute. Training to $1.6\times$ the original schedule cannot determine the unlimited-training limit. In summary, \method achieves better performance and compute efficiency within the tested practical training budget, with longer-term convergence left for future investigation.

\subsection{Robustness to Non-canonical Fields of View}
\label{sec:fov_robustness}

\begin{table}[t]
\centering
\caption{\textbf{Robustness to non-canonical fields of view.}
Foreground occupancy, organ occupancy, and AMOS 10\% Dice Score (\%) are reported.}
\label{tab:fov_robustness}
\setlength{\tabcolsep}{2.0mm}
\renewcommand{\arraystretch}{1.0}
\resizebox{1.00\linewidth}{!}{
\begin{tabular}{llccc}
\toprule
\textbf{Pretraining FOV}
& \textbf{Sampling Policy}
& \textbf{Foreground Occ. (\%) $\uparrow$}
& \textbf{Organ Occ. (\%) $\uparrow$}
& \textbf{AMOS 10\% Dice (\%) $\uparrow$} \\
\midrule
\multirow{2}{*}{Canonical}
& Random
& 55.83 & 19.79 & 73.34 \\

& \cellcolor{tableblue!80}\method-\lite
& \cellcolor{tableblue!80}\textbf{75.63} & \cellcolor{tableblue!80}\textbf{45.71} & \cellcolor{tableblue!80}\textbf{74.95} \\
\midrule
\multirow{3}{*}{Shifted/local FOV}
& Random
& 47.62 & 15.87 & 72.08 \\

& \cellcolor{tableblue!80}\method-\lite
& \cellcolor{tableblue!80}54.12 & \cellcolor{tableblue!80}25.13 & \cellcolor{tableblue!80}72.40 \\
& \method-I+D
& \textbf{60.94} & \textbf{31.13} & \textbf{73.71} \\
\bottomrule
\end{tabular}
}
\end{table}

To examine the spatial assumptions of \method-\lite, we construct a non-canonical field-of-view (FOV) stress test by shifting each foreground-cropped pretraining volume by 15--20\% in-plane or 8--12\% through-plane using crop-and-pad, moving anatomy toward the boundary and introducing partial truncation. Using VoCo~\cite{Wu_2024_CVPR}, we compare \textit{Random}, \method-\lite, and \method-I+D, which removes \textit{LocPrior} and \textit{Margin} while retaining informativeness and diversity.

Table~\ref{tab:fov_robustness} indicates that \method-\lite remains better than \textit{Random} under shifted/local FOVs, but its downstream advantage becomes smaller when the geometric assumptions are weakened. Removing the geometry terms improves robustness, with \method-I+D achieving the best occupancy and Dice under the shifted setting. The geometric prior is effective for foreground-normalized, approximately canonical volumes, but becomes less reliable under non-canonical fields of view. For strongly shifted or truncated scans, attenuating or removing the geometry terms may be preferable.

\subsection{Efficiency Analysis}
\label{app:efficiency}

To assess the computational overhead introduced by \method, we compare random sampling and \method-based sampling under each SSL backbone in Table~\ref{tab:efficiency_analysis}. All methods use their corresponding official preprocessing and pretraining settings, and all timing experiments are conducted on Intel(R) Data Center GPU Max 1550 GPUs. We report the one-time offline probe-bank construction time, probe-bank cache footprint, per-epoch pretraining time, total pretraining time, and the relative training overhead compared with \textit{Random} under the same backbone and training setup. The overhead is computed from total pretraining time, while offline construction time is reported separately. Since VoCo~\cite{Wu_2024_CVPR} and S2DC~\cite{Pan_2025_ICCV} employ the same VoCo-10k pretraining dataset, they share the same probe-bank construction pipeline and cached probe banks, leading to identical offline construction time for \method-\lite. Differences in sub-volume size lead to different \method-\feat probe-bank construction times for VoCo and S2DC. Additionally, the cache footprint is determined only by the number of pretraining volumes and the bank size, since both \method-\lite and \method-\feat ultimately cache the same type of 3D crop-center coordinates, independent of the SSL backbone.

Overall, \method remains lightweight and practical for large-scale pretraining. The total pretraining overhead is only 4.4\% for VoCo~\cite{Wu_2024_CVPR}, 2.1\% for S2DC~\cite{Pan_2025_ICCV}, and 2.2\% for MedGMAE~\cite{fu2026medgmae}. The cached probe banks require only 41.5 MB for VoCo-10k~\cite{Wu_2024_CVPR} and 21.8 MB for AbdomenAtlasMini~\cite{LI2024103285}, which is negligible compared with typical 3D medical pretraining datasets. Although feature extraction makes offline construction longer for \method-\feat than for \method-\lite, both variants have the same per-epoch and total pretraining time once the bank is cached. The above statistics confirm that \method preserves the original SSL pretraining pipeline, \textbf{adds only a small recurring cost, and introduces minimal storage overhead, making it practical for larger-scale deployments}.

\begin{table*}[t]
\centering
\caption{\textbf{Efficiency comparison between random sampling and \method.} For each SSL backbone, we report the probe-bank construction time, cache footprint, per-epoch pretraining time, total pretraining time, and relative overhead compared with \textit{Random} under the same training settings.}
\setlength{\tabcolsep}{1.5mm}
\resizebox{1.00\linewidth}{!}{
\begin{tabular}{lccccc}
\toprule
\textbf{Method} 
& \textbf{Offline Construction} 
& \textbf{Cache Footprint} 
& \textbf{Per-epoch Time} 
& \textbf{Total Pretraining Time} 
& \textbf{Overhead} \\
& \textbf{Time (h)} 
& \textbf{(MB)}
& \textbf{(min)} 
& \textbf{(GPU h)} 
& \textbf{vs. Random} \\
\midrule
VoCo~\cite{Wu_2024_CVPR} 
& -- 
& -- 
& 392.6
& 269.3 
& -- \\
\rowcolor{tableblue!80}
+ \method-\lite 
& 3.79  \
& 41.5
& 406.8
& 281.1
& +4.4\% \\
\rowcolor{tablepeach!70}
+ \method-\feat 
& 9.26 
& 41.5
& 406.8 
& 281.1 
& +4.4\% \\
\midrule
S2DC~\cite{Pan_2025_ICCV}
& -- 
& -- 
& 1089.9
& 753.3 
& -- \\
\rowcolor{tableblue!80}
+ \method-\lite 
& 3.79  
& 41.5
& 1137.4  
& 768.9 
& +2.1\% \\
\rowcolor{tablepeach!70}
+ \method-\feat 
& 9.57
& 41.5
& 1137.4 
& 768.9 
& +2.1\% \\
\midrule
MedGMAE~\cite{fu2026medgmae}
& -- 
& -- 
& 19.89 
& 662.9 
& -- \\
\rowcolor{tableblue!80}
+ \method-\lite 
& 1.84 
& 21.8
& 20.31
& 677.2
& +2.2\% \\
\rowcolor{tablepeach!70}
+ \method-\feat 
& 3.93
& 21.8
& 20.31 
& 677.2 
& +2.2\% \\
\bottomrule
\end{tabular}
}
\label{tab:efficiency_analysis}
\end{table*}

\section{Additional Analysis of \method}



\begin{figure*}[t]
\centering
\includegraphics[width=1.00\linewidth]{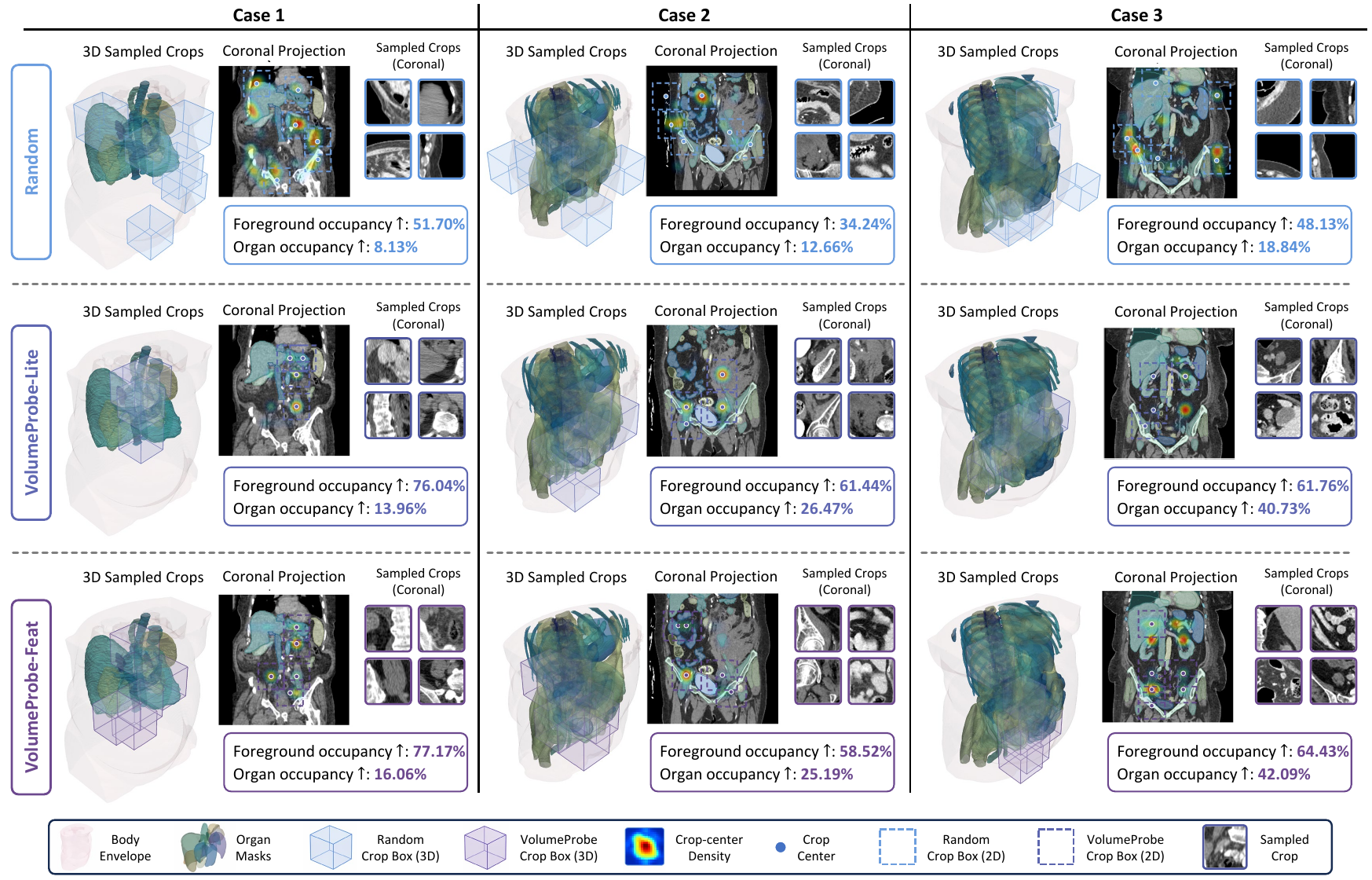}
\caption{\textbf{Additional cases for anatomy-aware sampling.}
We compare \textit{Random}, \method-\lite, and \method-\feat across three volumes.
For each case, we show 3D crop boxes overlaid with organ masks, coronal crop-center density, sampled crop centers, 2D crop boxes, and example cropped views. Both \method variants sample more anatomically valid and organ-rich regions than \textit{Random}, leading to higher foreground and organ occupancy ratios.}
\label{fig:volume_vis_APP}
\end{figure*}

\subsection{Additional Case Visualizations}
\label{app:case}
Fig.~\ref{fig:volume_vis_APP} provides additional case studies comparing \textit{Random}, \method-\lite, and \method-\feat. Across different volumes, \textit{Random} often samples boundary-dominated or weakly anatomical regions, while both \method variants place crops more consistently inside the body and around organ-rich areas. The improved foreground and organ occupancy ratios further imply that both \lite and \feat redirect the sub-crop budget toward anatomically valid observations.

\begin{figure*}[t]
\centering
\includegraphics[width=1.00\linewidth]{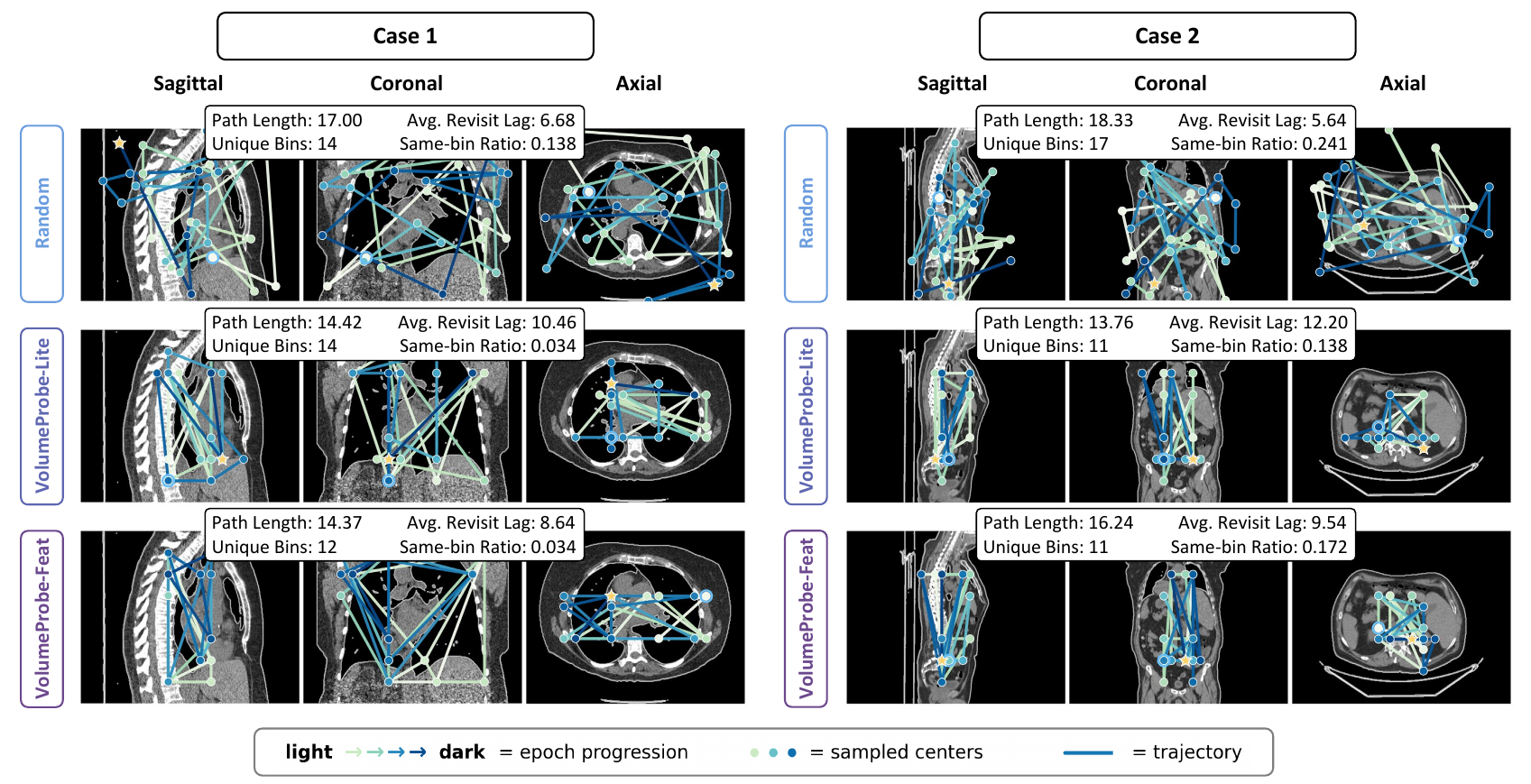}
\caption{\textbf{Cross-epoch trajectory analysis.} We visualize sampled crop centers across epochs for two representative cases under \textit{Random}, \method-\lite, and \method-\feat on sagittal, coronal, and axial views. The light-to-dark trajectory denotes epoch progression, and the inset reports path length, unique bins, average revisit lag, and same-bin ratio.}
\label{fig:cross_epoch_trajectory}
\end{figure*} 

\subsection{Cross-epoch Trajectory}
Fig.~\ref{fig:cross_epoch_trajectory} compares how different sampling strategies move across epochs under the same sub-volume cropping scheme. 
We summarize each trajectory using four metrics, i.e., path length measures the cumulative displacement between consecutive sampled centers, unique bins count the number of distinct coarse bins visited, average revisit lag measures the epoch interval before returning to a previously visited bin, and same-bin ratio quantifies the fraction of repeated visits within the same coarse bin. \textit{Random} often produces long but weakly structured trajectories with frequent local revisits, whereas \method-\lite and \method-\feat keep the trajectory more constrained to anatomically plausible regions. 
This is reflected by lower same-bin ratios and larger revisit lags, suggesting that the cached probe bank reduces redundant observations while preserving structured cross-epoch exploration. 
Notably, \method-\feat does not necessarily maximize path length, but tends to yield semantically more coherent trajectories within meaningful anatomical support.

\begin{figure*}[t]
\centering
\includegraphics[width=1.00\linewidth]{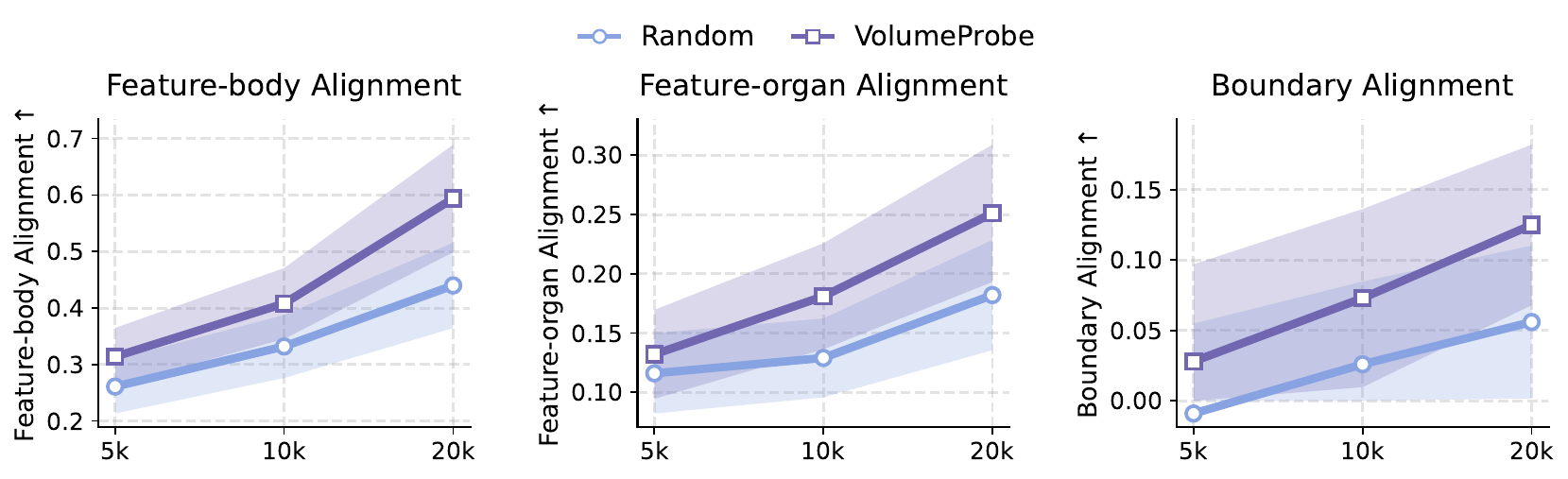}
\caption{\textbf{Quantitative analysis of structure-aligned feature responses.} We evaluate \textit{Random} and \method at 5k, 10k, and 20k early pretraining iterations using feature responses extracted from the same encoder stage. We report feature-body alignment, feature-organ alignment, and boundary alignment, with shaded regions indicating standard deviation across sampled sub-volumes. \method consistently achieves higher alignment scores than \textit{Random}, supporting the qualitative feature-map observations in Fig.~\ref{fig:learning}(c).}
\label{fig:feature_alignment}
\end{figure*}

\subsection{Additional Learning Relevance Analysis}
\label{app:additional_learning_relevance}
\textbf{Quantitative analysis of structure-aligned responses.}
To complement the qualitative feature-map visualization in Fig.~\ref{fig:learning}(c), we further quantify whether early feature responses align with anatomical structures. Specifically, we compute three metrics on BTCV~\cite{landman2015miccai} using channel-aggregated activation maps from the same encoder stage. Feature-body alignment and feature-organ alignment measure the fraction of activation mass falling inside the foreground body mask and organ-union mask, respectively, reflecting whether responses avoid background regions and focus on anatomically meaningful structures. Boundary alignment is measured by the Spearman correlation between activation responses and image-gradient magnitude within the body region, indicating whether features follow local structural transitions.

Fig.~\ref{fig:feature_alignment} reports the three alignment metrics across 5k, 10k, and 20k iterations, where \method consistently achieves higher scores than \textit{Random}. The improvements become more pronounced as pretraining proceeds, suggesting that \method encourages feature responses to concentrate more on valid body regions, organ-relevant areas, and local structural boundaries. These results provide quantitative support for the structure-aligned responses observed in Fig.~\ref{fig:learning}(c), mitigating the concern that the qualitative examples could be due to case selection.

\subsection{Analysis of the Feat Variant}

We further analyze why \method-\feat does not consistently outperform
\method-\lite despite using richer semantic features. We first clarify that MedicalNet~\cite{chen2019med3d} is used only for the diagnostic feature-level proxy $Q_2$ in the motivation study, whereas \method-\feat uses a frozen 3DINO-ViT~\cite{xu2025generalizable} encoder for offline probe-bank construction. As reported in the main results (Tables~\ref{tab:seg_cls} and~\ref{tab:reg}), \method-\feat consistently improves over \textit{Random} sampling, indicating that the frozen features provide useful semantic information, but it offers no consistent advantage over the lightweight \lite variant.

Several factors may contribute to this behavior. The frozen encoder is optimized for general semantic representation rather than local crop-utility estimation, while semantic invariance may suppress fine-grained intensity and structural variations directly captured by gradient and entropy cues. Moreover, the frozen feature space and coarse-bin prototypes are not optimized for crop utility and cannot adapt to volume-specific characteristics or evolving training stages. Consequently, the intrinsic cues in \lite appear more directly aligned with informative sub-volume selection under the current setting. This motivates \lite as our default variant, while task-adaptive or jointly learned semantic scoring remains an interesting direction for future work.


\clearpage
\section*{NeurIPS Paper Checklist}

\begin{enumerate}

\item {\bf Claims}
    \item[] Question: Do the main claims made in the abstract and introduction accurately reflect the paper's contributions and scope?
    \item[] Answer: \answerYes{} 
    \item[] Justification: The main claims presented in our abstract and introduction accurately reflect the contribution of our paper.
    \item[] Guidelines:
    \begin{itemize}
        \item The answer \answerNA{} means that the abstract and introduction do not include the claims made in the paper.
        \item The abstract and/or introduction should clearly state the claims made, including the contributions made in the paper and important assumptions and limitations. A \answerNo{} or \answerNA{} answer to this question will not be perceived well by the reviewers. 
        \item The claims made should match theoretical and experimental results, and reflect how much the results can be expected to generalize to other settings. 
        \item It is fine to include aspirational goals as motivation as long as it is clear that these goals are not attained by the paper. 
    \end{itemize}

\item {\bf Limitations}
    \item[] Question: Does the paper discuss the limitations of the work performed by the authors?
    \item[] Answer: \answerYes{} 
    \item[] Justification: We discuss the limitations of our work in Sec.~\ref{sec:conclusion}.
    \item[] Guidelines:
    \begin{itemize}
        \item The answer \answerNA{} means that the paper has no limitation while the answer \answerNo{} means that the paper has limitations, but those are not discussed in the paper. 
        \item The authors are encouraged to create a separate ``Limitations'' section in their paper.
        \item The paper should point out any strong assumptions and how robust the results are to violations of these assumptions (e.g., independence assumptions, noiseless settings, model well-specification, asymptotic approximations only holding locally). The authors should reflect on how these assumptions might be violated in practice and what the implications would be.
        \item The authors should reflect on the scope of the claims made, e.g., if the approach was only tested on a few datasets or with a few runs. In general, empirical results often depend on implicit assumptions, which should be articulated.
        \item The authors should reflect on the factors that influence the performance of the approach. For example, a facial recognition algorithm may perform poorly when image resolution is low or images are taken in low lighting. Or a speech-to-text system might not be used reliably to provide closed captions for online lectures because it fails to handle technical jargon.
        \item The authors should discuss the computational efficiency of the proposed algorithms and how they scale with dataset size.
        \item If applicable, the authors should discuss possible limitations of their approach to address problems of privacy and fairness.
        \item While the authors might fear that complete honesty about limitations might be used by reviewers as grounds for rejection, a worse outcome might be that reviewers discover limitations that aren't acknowledged in the paper. The authors should use their best judgment and recognize that individual actions in favor of transparency play an important role in developing norms that preserve the integrity of the community. Reviewers will be specifically instructed to not penalize honesty concerning limitations.
    \end{itemize}

\item {\bf Theory assumptions and proofs}
    \item[] Question: For each theoretical result, does the paper provide the full set of assumptions and a complete (and correct) proof?
    \item[] Answer: \answerNA{} 
    \item[] Justification: This work focuses on devising a plug-and-play, training-free module for sub-volume sampling in 3D medical SSL, which  does not contain theoretical results.
    \item[] Guidelines:
    \begin{itemize}
        \item The answer \answerNA{} means that the paper does not include theoretical results. 
        \item All the theorems, formulas, and proofs in the paper should be numbered and cross-referenced.
        \item All assumptions should be clearly stated or referenced in the statement of any theorems.
        \item The proofs can either appear in the main paper or the supplemental material, but if they appear in the supplemental material, the authors are encouraged to provide a short proof sketch to provide intuition. 
        \item Inversely, any informal proof provided in the core of the paper should be complemented by formal proofs provided in appendix or supplemental material.
        \item Theorems and Lemmas that the proof relies upon should be properly referenced. 
    \end{itemize}

    \item {\bf Experimental result reproducibility}
    \item[] Question: Does the paper fully disclose all the information needed to reproduce the main experimental results of the paper to the extent that it affects the main claims and/or conclusions of the paper (regardless of whether the code and data are provided or not)?
    \item[] Answer: \answerYes{} 
    \item[] Justification: We provide thorough experimental and implementation details, as well as hyperparameter settings in Appendix~\ref{app:imple}. The pretraining and downstream datasets (details in Appendix~\ref{app:data}) are all publicly available. 
    \item[] Guidelines:
    \begin{itemize}
        \item The answer \answerNA{} means that the paper does not include experiments.
        \item If the paper includes experiments, a \answerNo{} answer to this question will not be perceived well by the reviewers: Making the paper reproducible is important, regardless of whether the code and data are provided or not.
        \item If the contribution is a dataset and\slash or model, the authors should describe the steps taken to make their results reproducible or verifiable. 
        \item Depending on the contribution, reproducibility can be accomplished in various ways. For example, if the contribution is a novel architecture, describing the architecture fully might suffice, or if the contribution is a specific model and empirical evaluation, it may be necessary to either make it possible for others to replicate the model with the same dataset, or provide access to the model. In general. releasing code and data is often one good way to accomplish this, but reproducibility can also be provided via detailed instructions for how to replicate the results, access to a hosted model (e.g., in the case of a large language model), releasing of a model checkpoint, or other means that are appropriate to the research performed.
        \item While NeurIPS does not require releasing code, the conference does require all submissions to provide some reasonable avenue for reproducibility, which may depend on the nature of the contribution. For example
        \begin{enumerate}
            \item If the contribution is primarily a new algorithm, the paper should make it clear how to reproduce that algorithm.
            \item If the contribution is primarily a new model architecture, the paper should describe the architecture clearly and fully.
            \item If the contribution is a new model (e.g., a large language model), then there should either be a way to access this model for reproducing the results or a way to reproduce the model (e.g., with an open-source dataset or instructions for how to construct the dataset).
            \item We recognize that reproducibility may be tricky in some cases, in which case authors are welcome to describe the particular way they provide for reproducibility. In the case of closed-source models, it may be that access to the model is limited in some way (e.g., to registered users), but it should be possible for other researchers to have some path to reproducing or verifying the results.
        \end{enumerate}
    \end{itemize}

\item {\bf Open access to data and code}
    \item[] Question: Does the paper provide open access to the data and code, with sufficient instructions to faithfully reproduce the main experimental results, as described in supplemental material?
    \item[] Answer: \answerYes{} 
    \item[] Justification: The data and code used in our paper are publicly available.
    \item[] Guidelines:
    \begin{itemize}
        \item The answer \answerNA{} means that paper does not include experiments requiring code.
        \item Please see the NeurIPS code and data submission guidelines (\url{https://neurips.cc/public/guides/CodeSubmissionPolicy}) for more details.
        \item While we encourage the release of code and data, we understand that this might not be possible, so \answerNo{} is an acceptable answer. Papers cannot be rejected simply for not including code, unless this is central to the contribution (e.g., for a new open-source benchmark).
        \item The instructions should contain the exact command and environment needed to run to reproduce the results. See the NeurIPS code and data submission guidelines (\url{https://neurips.cc/public/guides/CodeSubmissionPolicy}) for more details.
        \item The authors should provide instructions on data access and preparation, including how to access the raw data, preprocessed data, intermediate data, and generated data, etc.
        \item The authors should provide scripts to reproduce all experimental results for the new proposed method and baselines. If only a subset of experiments are reproducible, they should state which ones are omitted from the script and why.
        \item At submission time, to preserve anonymity, the authors should release anonymized versions (if applicable).
        \item Providing as much information as possible in supplemental material (appended to the paper) is recommended, but including URLs to data and code is permitted.
    \end{itemize}

\item {\bf Experimental setting/details}
    \item[] Question: Does the paper specify all the training and test details (e.g., data splits, hyperparameters, how they were chosen, type of optimizer) necessary to understand the results?
    \item[] Answer: \answerYes{} 
    \item[] Justification: All pretraining and downstream fine-tuning details are thoroughly described in Sec.~\ref{sec:exp_set} and Appendix~\ref{app:imple}, as well as ablations for obtaining  hyperparameters in Appendix~\ref{app:ablation}.
    \item[] Guidelines:
    \begin{itemize}
        \item The answer \answerNA{} means that the paper does not include experiments.
        \item The experimental setting should be presented in the core of the paper to a level of detail that is necessary to appreciate the results and make sense of them.
        \item The full details can be provided either with the code, in appendix, or as supplemental material.
    \end{itemize}

\item {\bf Experiment statistical significance}
    \item[] Question: Does the paper report error bars suitably and correctly defined or other appropriate information about the statistical significance of the experiments?
    \item[] Answer: \answerYes{} 
    \item[] Justification: We report mean and standard deviation over three independent runs for the downstream segmentation, classification, and registration results in Appendix~\ref{app:multi-run_robustness}. 
    \item[] Guidelines:
    \begin{itemize}
        \item The answer \answerNA{} means that the paper does not include experiments.
        \item The authors should answer \answerYes{} if the results are accompanied by error bars, confidence intervals, or statistical significance tests, at least for the experiments that support the main claims of the paper.
        \item The factors of variability that the error bars are capturing should be clearly stated (for example, train/test split, initialization, random drawing of some parameter, or overall run with given experimental conditions).
        \item The method for calculating the error bars should be explained (closed form formula, call to a library function, bootstrap, etc.)
        \item The assumptions made should be given (e.g., Normally distributed errors).
        \item It should be clear whether the error bar is the standard deviation or the standard error of the mean.
        \item It is OK to report 1-sigma error bars, but one should state it. The authors should preferably report a 2-sigma error bar than state that they have a 96\% CI, if the hypothesis of Normality of errors is not verified.
        \item For asymmetric distributions, the authors should be careful not to show in tables or figures symmetric error bars that would yield results that are out of range (e.g., negative error rates).
        \item If error bars are reported in tables or plots, the authors should explain in the text how they were calculated and reference the corresponding figures or tables in the text.
    \end{itemize}

\item {\bf Experiments compute resources}
    \item[] Question: For each experiment, does the paper provide sufficient information on the computer resources (type of compute workers, memory, time of execution) needed to reproduce the experiments?
    \item[] Answer: \answerNo{} 
    \item[] Justification: Given the number of experiments conducted in this work, we are unable to give an exhaustive list of compute used for each experiment. However, we provide batch sizes and hyperparameters for all pretraining and fine-tuning experiments in Appendix~\ref{app:imple}, as well as pretraining time in Appendix~\ref{app:efficiency}.
    \item[] Guidelines:
    \begin{itemize}
        \item The answer \answerNA{} means that the paper does not include experiments.
        \item The paper should indicate the type of compute workers CPU or GPU, internal cluster, or cloud provider, including relevant memory and storage.
        \item The paper should provide the amount of compute required for each of the individual experimental runs as well as estimate the total compute. 
        \item The paper should disclose whether the full research project required more compute than the experiments reported in the paper (e.g., preliminary or failed experiments that didn't make it into the paper). 
    \end{itemize}
    
\item {\bf Code of ethics}
    \item[] Question: Does the research conducted in the paper conform, in every respect, with the NeurIPS Code of Ethics \url{https://neurips.cc/public/EthicsGuidelines}?
    \item[] Answer: \answerYes{} 
    \item[] Justification: We have read and understood the NeurIPS Code of Ethics and have done our best to conform.
    \item[] Guidelines:
    \begin{itemize}
        \item The answer \answerNA{} means that the authors have not reviewed the NeurIPS Code of Ethics.
        \item If the authors answer \answerNo, they should explain the special circumstances that require a deviation from the Code of Ethics.
        \item The authors should make sure to preserve anonymity (e.g., if there is a special consideration due to laws or regulations in their jurisdiction).
    \end{itemize}

\item {\bf Broader impacts}
    \item[] Question: Does the paper discuss both potential positive societal impacts and negative societal impacts of the work performed?
    \item[] Answer: \answerNA{} 
    \item[] Justification: There is no potential societal impact of our work.
    \item[] Guidelines:
    \begin{itemize}
        \item The answer \answerNA{} means that there is no societal impact of the work performed.
        \item If the authors answer \answerNA{} or \answerNo, they should explain why their work has no societal impact or why the paper does not address societal impact.
        \item Examples of negative societal impacts include potential malicious or unintended uses (e.g., disinformation, generating fake profiles, surveillance), fairness considerations (e.g., deployment of technologies that could make decisions that unfairly impact specific groups), privacy considerations, and security considerations.
        \item The conference expects that many papers will be foundational research and not tied to particular applications, let alone deployments. However, if there is a direct path to any negative applications, the authors should point it out. For example, it is legitimate to point out that an improvement in the quality of generative models could be used to generate Deepfakes for disinformation. On the other hand, it is not needed to point out that a generic algorithm for optimizing neural networks could enable people to train models that generate Deepfakes faster.
        \item The authors should consider possible harms that could arise when the technology is being used as intended and functioning correctly, harms that could arise when the technology is being used as intended but gives incorrect results, and harms following from (intentional or unintentional) misuse of the technology.
        \item If there are negative societal impacts, the authors could also discuss possible mitigation strategies (e.g., gated release of models, providing defenses in addition to attacks, mechanisms for monitoring misuse, mechanisms to monitor how a system learns from feedback over time, improving the efficiency and accessibility of ML).
    \end{itemize}
    
\item {\bf Safeguards}
    \item[] Question: Does the paper describe safeguards that have been put in place for responsible release of data or models that have a high risk for misuse (e.g., pre-trained language models, image generators, or scraped datasets)?
    \item[] Answer: \answerNA{} 
    \item[] Justification: Our experiments do not involve this aspect.
    \item[] Guidelines:
    \begin{itemize}
        \item The answer \answerNA{} means that the paper poses no such risks.
        \item Released models that have a high risk for misuse or dual-use should be released with necessary safeguards to allow for controlled use of the model, for example by requiring that users adhere to usage guidelines or restrictions to access the model or implementing safety filters. 
        \item Datasets that have been scraped from the Internet could pose safety risks. The authors should describe how they avoided releasing unsafe images.
        \item We recognize that providing effective safeguards is challenging, and many papers do not require this, but we encourage authors to take this into account and make a best faith effort.
    \end{itemize}

\item {\bf Licenses for existing assets}
    \item[] Question: Are the creators or original owners of assets (e.g., code, data, models), used in the paper, properly credited and are the license and terms of use explicitly mentioned and properly respected?
    \item[] Answer: \answerYes{} 
    \item[] Justification: We cite the original paper that produced the code package or dataset.
    \item[] Guidelines:
    \begin{itemize}
        \item The answer \answerNA{} means that the paper does not use existing assets.
        \item The authors should cite the original paper that produced the code package or dataset.
        \item The authors should state which version of the asset is used and, if possible, include a URL.
        \item The name of the license (e.g., CC-BY 4.0) should be included for each asset.
        \item For scraped data from a particular source (e.g., website), the copyright and terms of service of that source should be provided.
        \item If assets are released, the license, copyright information, and terms of use in the package should be provided. For popular datasets, \url{paperswithcode.com/datasets} has curated licenses for some datasets. Their licensing guide can help determine the license of a dataset.
        \item For existing datasets that are re-packaged, both the original license and the license of the derived asset (if it has changed) should be provided.
        \item If this information is not available online, the authors are encouraged to reach out to the asset's creators.
    \end{itemize}

\item {\bf New assets}
    \item[] Question: Are new assets introduced in the paper well documented and is the documentation provided alongside the assets?
    \item[] Answer: \answerNA{} 
    \item[] Justification: The paper does not release new assets.
    \item[] Guidelines:
    \begin{itemize}
        \item The answer \answerNA{} means that the paper does not release new assets.
        \item Researchers should communicate the details of the dataset\slash code\slash model as part of their submissions via structured templates. This includes details about training, license, limitations, etc. 
        \item The paper should discuss whether and how consent was obtained from people whose asset is used.
        \item At submission time, remember to anonymize your assets (if applicable). You can either create an anonymized URL or include an anonymized zip file.
    \end{itemize}

\item {\bf Crowdsourcing and research with human subjects}
    \item[] Question: For crowdsourcing experiments and research with human subjects, does the paper include the full text of instructions given to participants and screenshots, if applicable, as well as details about compensation (if any)? 
    \item[] Answer: \answerNA{} 
    \item[] Justification: This paper does not involve this aspect.
    \item[] Guidelines:
    \begin{itemize}
        \item The answer \answerNA{} means that the paper does not involve crowdsourcing nor research with human subjects.
        \item Including this information in the supplemental material is fine, but if the main contribution of the paper involves human subjects, then as much detail as possible should be included in the main paper. 
        \item According to the NeurIPS Code of Ethics, workers involved in data collection, curation, or other labor should be paid at least the minimum wage in the country of the data collector. 
    \end{itemize}

\item {\bf Institutional review board (IRB) approvals or equivalent for research with human subjects}
    \item[] Question: Does the paper describe potential risks incurred by study participants, whether such risks were disclosed to the subjects, and whether Institutional Review Board (IRB) approvals (or an equivalent approval/review based on the requirements of your country or institution) were obtained?
    \item[] Answer: \answerNA{} 
    \item[] Justification: This paper does not involve this aspect.
    \item[] Guidelines:
    \begin{itemize}
        \item The answer \answerNA{} means that the paper does not involve crowdsourcing nor research with human subjects.
        \item Depending on the country in which research is conducted, IRB approval (or equivalent) may be required for any human subjects research. If you obtained IRB approval, you should clearly state this in the paper. 
        \item We recognize that the procedures for this may vary significantly between institutions and locations, and we expect authors to adhere to the NeurIPS Code of Ethics and the guidelines for their institution. 
        \item For initial submissions, do not include any information that would break anonymity (if applicable), such as the institution conducting the review.
    \end{itemize}

\item {\bf Declaration of LLM usage}
    \item[] Question: Does the paper describe the usage of LLMs if it is an important, original, or non-standard component of the core methods in this research? Note that if the LLM is used only for writing, editing, or formatting purposes and does \emph{not} impact the core methodology, scientific rigor, or originality of the research, declaration is not required.
    \item[] Answer: \answerNA{} 
    \item[] Justification: This work does not incorporate any LLMs as part of the core methodology. LLMs were used only for writing and editing suggestions in this paper.
    \item[] Guidelines:
    \begin{itemize}
        \item The answer \answerNA{} means that the core method development in this research does not involve LLMs as any important, original, or non-standard components.
        \item Please refer to our LLM policy in the NeurIPS handbook for what should or should not be described.
    \end{itemize}

\end{enumerate}

\end{document}